\documentclass{uai2022} 

\usepackage{lmodern}
\usepackage[english]{babel}
\usepackage{latexsym}
\usepackage{amsmath}
\usepackage{mathrsfs}
\usepackage{amssymb}
\usepackage{mathtools}
\usepackage{bm}
\usepackage{datetime}
\usepackage[table,xcdraw]{xcolor}
\usepackage{accents}
\usepackage{tikz}
\usepackage{listings}
\usepackage{mdframed}
\usepackage[algoruled, linesnumbered]{algorithm2e} 
\usepackage{xr}

\usepackage{dsfont}
\usepackage{color}
\usepackage{colortbl}
\usepackage{pifont}
\usepackage{caption}
\usepackage{microtype} 
\usepackage{float}
\usepackage{xfrac} 
\usepackage{xspace}
\usepackage{booktabs}
\usepackage{blkarray}
\usepackage{graphicx}
\graphicspath{{plot/}}
\usepackage{subcaption}

\usepackage[textsize=tiny,
disable
]{todonotes}
\newcommand{\todoc}[1]{\todo[color=orange!20!white]{Csaba: #1}}

\newcommand{\red}[1]{\textcolor{red}{#1}}
\newcommand{\blue}[1]{\textcolor{blue}{#1}}
\newcommand{\teal}[1]{\textcolor{teal}{#1}}

\usepackage{hyperref}
\hypersetup{
    unicode=false,          
    pdftoolbar=true,        
    pdfmenubar=true,        
    pdffitwindow=false,     
    pdfstartview={FitH},    
    pdftitle={XXXXX},    
    pdfauthor={XXX},     
    pdfsubject={Bandits, Reinforcement Learning},   
    pdfcreator={Creator},   
    pdfproducer={Producer}, 
    pdfkeywords={bandits} {reinforcement learning} {policy gradient}, 
    pdfnewwindow=true,      
    colorlinks=true,       
    linkcolor=blue,          
    citecolor=blue,        
    filecolor=magenta,      
    urlcolor=cyan           
}
\usepackage{amsthm}
\usepackage{times}
\usepackage{nicefrac}
\usepackage{wrapfig}
\usepackage{pgfplots}
\usepackage[capitalize]{cleveref}
\usepackage{thm-restate}

\theoremstyle{plain}
\newtheorem{theorem}{Theorem}[section]

\newtheorem{lemma}[theorem]{Lemma}

\theoremstyle{definition}

\newtheorem{assumption}[theorem]{Assumption}

\theoremstyle{remark}

\usepackage[round]{natbib} 
\makeatletter
\def\thm@space@setup{\thm@preskip=0pt
\thm@postskip=0pt}
\makeatother

\mdfdefinestyle{mdframedthmbox}{%
	leftmargin=.0\textwidth,
	rightmargin=.0\textwidth,%
	innertopmargin=0.75em,
	innerleftmargin=.5em,
	innerrightmargin=.5em,
}

\newenvironment{thmbox}
	{%
		\begin{mdframed}[style=mdframedthmbox]%
	}{%
		\end{mdframed}%
	}

\newcommand{\myquote}[1]{\null~\\{\null\hspace{.05\textwidth}\begin{minipage}[t]{.90\textwidth} #1 \end{minipage}}}

\crefname{algline}{Line}{Line}
\crefname{algline}{Line}{Line}

\newcommand{\E}{\mathbb{E}}

\newcommand{\PP}{\mathbb{P}}

\newcommand{\norm}[1]{\left\|#1\right\|}
\newcommand{\norminf}[1]{\left\|#1\right\|_{\infty}}
\newcommand{\indnorm}[2]{\|#1\|_{#2}}

\newcommand{\R}{\mathbb{R}}

\newcommand{\cA}{\mathcal{A}}

\newcommand{\cC}{\mathcal{C}}

\newcommand{\cM}{\mathcal{M}}
\newcommand{\cP}{\mathcal{P}}

\newcommand{\cR}{\mathcal{R}}
\newcommand{\cS}{\mathcal{S}}

\renewcommand{\epsilon}{\varepsilon}
\newcommand{\tvarepsilon}{\tilde{\varepsilon}}

\DeclareMathOperator*{\argmin}{arg\ min}
\DeclareMathOperator*{\argmax}{arg\ max}

\theoremstyle{plain}

\newcommand{\Alg}{\mathcal{A}}

\newcommand{\vals}[1]{V^{#1}}

\newcommand{\const}[1]{V_c^{#1}(\rho)}
\newcommand{\consthat}[1]{\hat{V}_c^{#1}(\rho)}

\newcommand{\consts}[1]{V_c^{#1}}

\newcommand{\reward}[1]{V_r^{#1}(\rho)}

\newcommand{\rewards}[1]{V_r^{#1}}

\newcommand{\lag}[2]{V_{l}^{#1,#2}(\rho)}

\newcommand{\lagqhat}[1]{\hat{Q}_{l}^{#1}}

\newcommand{\lagahat}[1]{\hat{A}_{l}^{#1}}
\newcommand{\lagatilde}[1]{\tilde{A}_{l}^{#1}}

\newcommand{\cI}{\mathcal{I}}

\newcommand{\rewardq}[1]{Q_{r}^{#1}}
\newcommand{\rewardqhat}[1]{\hat{Q}_{r}^{#1}}

\newcommand{\rewardqhats}[1]{\hat{Q}_{r}^{#1}(s,\cdot)}

\newcommand{\constq}[1]{Q_{c}^{#1}}
\newcommand{\constqhat}[1]{\hat{Q}_{c}^{#1}}

\newcommand{\constqhats}[1]{\hat{Q}_{c}^{#1}(s,\cdot)}

\newcommand{\piopt}{\pi^*}
\newcommand{\pit}{\pi_{t}}

\newcommand{\pitil}{\tilde{\pi}}
\newcommand{\tautil}{\tilde{\tau}}
\newcommand{\pip}{\pi^{\prime}}
\newcommand{\pitt}{\pi_{t+1}}
\newcommand{\lambdat}{\lambda_{t}}
\newcommand{\lambdatt}{\lambda_{t+1}}
\newcommand{\transpose}{^\mathsf{\scriptscriptstyle T}}

\newcommand{\epsb}{\varepsilon_{\text{\tiny{b}}}}
\newcommand{\epss}{\varepsilon_{\text{\tiny{s}}}}

\def\Alg/{{CBP}}
\newcommand{\Tau}{\mathrm{T}}
\newcommand{\thetahat}{\hat{\theta}}

\def\cAlg/{\red{CBP}}
\def\cGDA/{\blue{GDA}}
\def\cCRPO/{\teal{CRPO}}

\title{Towards Painless Policy Optimization for Constrained MDPs}

\author[1$^*$]{\href{mailto:<arushi.jain@mail.mcgill.ca>}{Arushi Jain}}
\author[2\thanks{Equal contribution}]{Sharan Vaswani}
\author[3]{Reza Babanezhad}
\author[4,5]{Csaba Szepesv\'ari}
\author[1,5]{Doina Precup}

\affil[1]{%
    Mila, McGill University
}
\affil[2]{%
   Simon Fraser University
}
\affil[3]{%
   SAIT AI Lab, Montreal
  }
\affil[4]{%
   Amii, University of Alberta 
   }
\affil[5]{%
DeepMind
}

\begin{document}
\maketitle

\begin{abstract}
We study policy optimization in an infinite horizon, $\gamma$-discounted constrained Markov decision process (CMDP). Our objective is to return a policy that achieves large expected reward with a small constraint violation. We consider the online setting with linear function approximation and assume global access to the corresponding features. We propose a generic primal-dual framework that allows us to bound the reward sub-optimality and constraint violation for arbitrary algorithms in terms of their primal and dual regret on online linear optimization problems. We instantiate this framework to use coin-betting algorithms and propose the \textbf{Coin Betting Politex (CBP)} algorithm. Assuming that the action-value functions are $\varepsilon_{\text{\tiny{b}}}$-close to the span of the $d$-dimensional state-action features and no sampling errors, we prove that $T$ iterations of CBP result in an $O\left(\frac{1}{(1 - \gamma)^3 \sqrt{T}} + \frac{\varepsilon_{\text{\tiny{b}}} \sqrt{d}}{(1 - \gamma)^2} \right)$ reward sub-optimality and an $O\left(\frac{1}{(1 - \gamma)^2 \sqrt{T}} + \frac{\varepsilon_{\text{\tiny{b}}} \sqrt{d}}{1 - \gamma} \right)$ constraint violation. Importantly, unlike gradient descent-ascent and other recent methods, CBP does not require extensive hyperparameter tuning. Via experiments on synthetic and Cartpole environments, we demonstrate the effectiveness and robustness of CBP.
\end{abstract}
\section{Introduction}
\label{sec:Introduction}
Popular reinforcement learning (RL) algorithms focus on optimizing an unconstrained objective, and have found applications in games such as Atari~\citep{mnih2015human} or Go~\citep{silver2016mastering}, robot manipulation tasks~\citep{tan2018sim,zeng2020tossingbot} or clinical trials~\citep{schaefer2005modeling}. However, many applications require the planning agent to satisfy constraints -- for example, in wireless sensor networks~\citep{buratti2009overview, julian2002qos} there is a constraint on  average power consumption of a deployed policy. Similarly, in safe RL, the policy is constrained to only visit certain states while exploring in physical systems ~\citep{moldovan2012safe,ono2015chance,fisac2018general}. The constrained Markov decision process (CMDP)~\citep{altman1999constrained} is a natural framework to model long-term constraints that need to be satisfied by a policy. The typical objective for CMDPs is to maximize the cumulative reward (similar to unconstrained MDPs), while (approximately) satisfying the constraint.

We focus on a well-studied problem in CMDPs -- return an approximately feasible policy (that is allowed to violate the constraints by a small amount), while (approximately) maximizing the cumulative reward. The past literature on this topic considered two approaches. The first approach is \emph{primal-only algorithms}, where constraints are (approximately) enforced without directly relying on introducing a Lagrangian formulation~\citep{achiam2017constrained, chow2018lyapunov,dalal2018safe,liu2020ipo,xu2021crpo}.
Of these methods, only the recent work of~\citet{xu2021crpo} guarantees global convergence to the optimal feasible policy in both the tabular and function approximation settings. 

The second approach in CMDPs is to form the Lagrangian, and solve the resulting saddle-point problem using \emph{primal-dual algorithms}~\citep{altman1999constrained,borkar2005actor,bhatnagar2012online,borkar2014risk,tessler2018reward,liang2018accelerated, paternain2019constrained, yu2019convergent, ding2021provably, ding2020natural, stooke2020responsive}. Such approaches update both the policy parameters (primal variables), while updating the Lagrange multipliers (dual variables). Of these methods,~\citet{tessler2018reward} prove a local convergence guarantee, while~\citet{paternain2019constrained} prove that their proposed algorithm will converge to a neighbourhood of the optimal policy. More recently, \citet{ding2020natural} proposed to use natural policy gradient updates~\citep{kakade2001natural} for changing the policy parameters
while using gradient descent to update the dual variables. They prove that this primal-dual algorithm converges to the optimal policy in both the tabular and the function approximation settings.   

\begin{figure}[t]
		\begin{subfigure}[b]{0.48\linewidth}
			\centering
			\captionsetup{justification=centering}
			\includegraphics[width=0.98\textwidth]{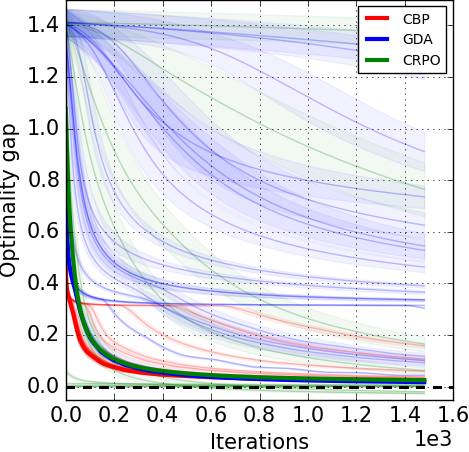}
			\caption[]{\small{Optimality gap\\ (OG)}}
		\end{subfigure}
		\begin{subfigure}[b]{0.49\linewidth}
			\centering
			\captionsetup{justification=centering}
			\includegraphics[width=0.99\textwidth]{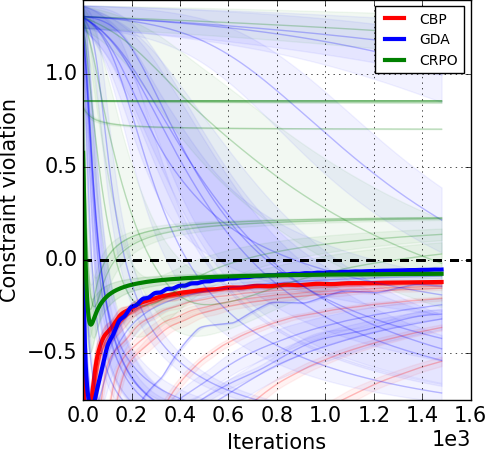}
			\caption[]{\small{Constraint violation (CV)}}
		\end{subfigure}
		\caption[]
        {\textbf{Hyperparameter sensitivity:} Optimality gap and constraint violation (averaged across $5$ runs) for different hyperparameters for \textbf{\cGDA/}~\citep{ding2020natural}, \textbf{\cCRPO/}~\citep{xu2021crpo}
        and the proposed algorithm \textbf{\cAlg/} on a gridworld environment with access to the true CMDP. The dark lines show the performance of the best hyperparameter, while the lighter-shade lines represent results while using other hyperparameters. Both GDA and CRPO exhibit large variations 
        in their performance, while \Alg/ is more robust. See~\cref{sec:Experiments} for details.}          
        \label{fig:sensitivity-intro}
\end{figure}

Although there is no lack in algorithms designed for CMDPs, \emph{these algorithms are often highly sensitive to the choice of their hyperparameters}. For example,~\cref{fig:sensitivity-intro} demonstrates the effect of varying the hyperparameters for two provably efficient algorithms, the primal-dual natural-policy ascent, gradient descent method (in short, GDA) of \citet{ding2020natural} and the primal-only CRPO method of~\citet{xu2021crpo} on a synthetic tabular environment. 
While one can find hyperparameters that control the worst-case performance of either GDA or CRPO, 
such choices result in a poor empirical performance on individual instances, a feature that GDA and CRPO share with
unconstrained MDP policy optimization algorithms, such as Politex~\citep{abbasi2019politex}, 
or natural policy gradient~\citep{kakade2001natural}.

\paragraph{CONTRIBUTIONS:} \textit{Designing robust policy optimization algorithms that require minimal hyperparameter tuning is our main motivation, and towards this, we make the following contributions.}

\textbf{Generic Primal-Dual Framework}: In~\cref{sec:framework}, we cast the problem of planning in discounted infinite horizon CMDPs to a generic primal-dual framework. In particular, we prove that any algorithm that can control (i) the primal and dual regret for specific online linear optimization problems and (ii) the errors due to function approximation and sampling, will (approximately) maximize the cumulative discounted reward while (approximately) minimizing the constraint violation (\cref{thm:generic-bound}). Importantly, this result holds for any CMDP and is independent of how the policies or value functions are represented.

\textbf{Instantiating the Framework}: In~\cref{sec:framework-instantiation}, we instantiate the framework using two algorithms from the online linear optimization literature -- Gradient Descent Ascent (GDA) (\cref{sec:gda}) and Coin-Betting (CB) (\cref{sec:cb}). While GDA requires setting the hyperparameters to specific problem-dependent constants, CB is more robust to hyperparameter tuning (see \cref{fig:sensitivity-intro}). In the simpler tabular setting, the approximation errors can be easily controlled and we use~\cref{thm:generic-bound} in conjunction with existing regret bounds to prove that the average optimality gap (difference in the cumulative reward of achieved policy and the optimal policy) and the average constraint violation decrease at an $O\left(\nicefrac{1}{\sqrt{T}}\right)$ rate (\cref{cor:gda,cor:cb}). 

\textbf{Handling Linear Function Approximation}: In~\cref{sec:putting-together}, we assume global access to a $d$-dimensional feature map $\Phi : \cS \times \cA \rightarrow \R^d$, and that the action-value functions for any policy are $\epsb$-close to the span of these features. With this assumption, we prove that it is possible to control the approximation errors for each state-action pair. Subsequently, in~\cref{sec:lfa-algorithm}, we use the robust coin-betting algorithms to instantiate the primal-dual framework in the linear function approximation setting and propose the \emph{Coin-Betting Politex} (\Alg/) algorithm. Ignoring sampling errors, in~\cref{sec:linear-bound}, we prove that the average optimality gap for \Alg/ scales as $O \left(\frac{1}{(1 - \gamma)^3 \, \sqrt{T}} + \frac{\epsb \sqrt{d}}{(1 - \gamma)^2} \right)$, while the average constraint violation is $O \left(\frac{1}{(1 - \gamma)^2 \, \sqrt{T}} + \frac{\epsb \sqrt{d}}{(1 - \gamma)} \right)$. With linear function approximation, the average constraint violation for the algorithm of~\citet{ding2020natural} decreases at a worse $O\left(\nicefrac{1}{T^{1/4}} \right)$ rate. On the other hand, the CRPO algorithm of~\citet{xu2021crpo} results in an $O\left(\nicefrac{1}{\sqrt{T}} \right)$ bound for both the average suboptimality and constraint violation. However, both algorithms can amplify the function approximation errors to large, potentially unbounded values. Importantly, both algorithms require typically unknown quantities which impedes their practical use. 

\textbf{Experimental Evaluation}: In~\cref{sec:Experiments}, we first describe some practical considerations when implementing \Alg/. We then evaluate \Alg/ and compare its empirical performance to the algorithms of~\citet{ding2020natural,xu2021crpo}. Our experiments on synthetic tabular environment and the Cartpole environment with linear  function approximation demonstrate the consistent effectiveness and robustness of \Alg/. 


\section{Problem Formulation}
\label{sec:problem-formulation}
We consider an infinite-horizon discounted constrained Markov decision process (CMDP)~\citep{altman1999constrained} defined by the tuple $\langle \cS, \cA, \cP, r, c, b, \rho, \gamma \rangle$ where $\cS$ is the countable set of states, $\cA$ is the countable action set, $\cP : \cS \times \cA \rightarrow \Delta_\cS$ is the transition probability function, $\Delta_\cS$ is the $\cS$-dimensional probability simplex, $\rho \in \Delta_{\cS}$ is the initial distribution of states and $\gamma \in [0, 1)$ is the discount factor. The primary reward to be maximized is denoted by $r : \cS \times \cA \rightarrow [0,1]$. For each state $s$, we define the reward value function w.r.t. the policy $\pi : \cS \rightarrow \Delta_{\cA}$ as $V_r^\pi(s)=\mathbb{E}_{\pi,\cP} \Big[\sum_{t=0}^\infty \gamma^t r(s_t, a_t)| s_0=s\Big]$ where $ a_t\sim\pi( \cdot| s_t),$ and $s_{t+1}\sim\cP(\cdot | s_t, a_t)$ and $\Delta_{\cA}$ is $\cA$-dimensional simplex. The expected discounted return or \emph{reward value} of a policy $\pi$ is defined as $\reward{\pi} = \mathbb{E}_{s_0\sim \rho} \Big[V_r^\pi(s_0)\Big]$. Similarly, the constraint reward is denoted by $c: \cS \times \cA \rightarrow [0,1]$ and the \emph{constraint reward value} for $\pi$ by $\const{\pi}$. 
For each $(s,a)$ under policy $\pi$, the reward action-value function is defined as $\rewardq{\pi}: \cS \times \cA \rightarrow \R$ s.t. $\rewardq{\pi}(s,a) = r(s,a)+\gamma\E_{s'\sim\cP(\cdot|s,a)}[V_{r}^{\pi}(s')]$ and satisfies the relation: $V_r^\pi(s) = \langle \pi(\cdot | s), \rewardq{\pi}(s,\cdot) \rangle(:=\E_{a\sim \pi(\cdot|s)}[\rewardq{\pi}(s,a)])$. 
We define $\constq{\pi}$ analogously. The agent's objective is to return a policy $\pi$ that maximizes $\reward{\pi}$, while ensuring that $\const{\pi} \geq b$. Formally,  
\begin{align}
\max_{\pi} \reward{\pi} \quad \text{s.t.} \quad  \const{\pi} \geq b.
\label{eq:objective}
\end{align}
Throughout, we will assume the existence of a feasible policy (i.e., one with $ \const{\pi} \geq b$), and denote the optimal feasible policy by $\piopt$. 
Due to sampling and other errors, we will aim for finding a policy $\pi$  such that,
\begin{align}
\reward{\pi} & \geq \reward{\piopt} - \epsilon  \quad \text{s.t.} \quad  \const{\pi} \geq b - \epsilon
\label{eq:relaxed-objective}
\end{align}
with some $\epsilon>0$.
In the next section, we specify a generic primal-dual framework solving the problem in~\cref{eq:relaxed-objective}.

\section{Primal-Dual Framework}
\label{sec:framework}
By Lagrangian duality, $\pi^*$ is a solution to \cref{eq:objective} if and only if for some $\lambda^*\ge 0$,
$(\pi^*,\lambda^*)$ solves the saddle-point problem  
\begin{align}
\max_{\pi} \min_{\lambda \geq 0} \reward{\pi} + \lambda [\const{\pi} - b]\,.  
\label{eq:objective-saddle}
\end{align}
Here, $\lambda \in \R$ is the Lagrange multiplier for the constraint.

We will solve the above primal-dual saddle-point problem iteratively, by alternatively updating the policy (primal variable) and the Lagrange multiplier (dual variable). If $T$ is the total number of iterations, we define $\pit$ and $\lambdat$ to be the primal and dual iterates for $t \in [T]:=\{1,\dots,T\}$. Updating the $(\pit, \lambdat)$ variables will require estimating the action-value functions. We define $\rewardqhat{t} := \rewardqhat{\pit}$ and $\constqhat{t} := \constqhat{\pit}$ as the \emph{estimated} action-value functions corresponding to the policy $\pit$. We also define $\hat{V}_c^{\pi}(s) := \langle \pi(\cdot|s),\constqhat{t}(s,\cdot)\rangle$, $\reward{t}: = \reward{\pit}$ and $\const{t}:=\const{\pit}$.
In this section, we assume that  $\norm{\rewardq{t} - \rewardqhat{t}}_{\infty} \leq \tvarepsilon$ and $\norm{\constq{t} - \constqhat{t}}_{\infty} \leq \tvarepsilon$. 

Given a generic primal-dual algorithm, our task is to characterize its performance in terms of its cumulative reward and constraint violation. Specifically, for a sequence of policies $\{\pi_0, \pi_1, \ldots, \pi_{T-1}\}$ and Lagrange multipliers $\{\lambda_0, \lambda_1, \ldots, \lambda_{T-1}\}$ generated by an algorithm, we define the \emph{average optimality gap} (\texttt{OG}) and the \emph{average constraint violation} (\texttt{CV}) as,
\begin{align}
&\text{Average optimality gap (\texttt{OG})} := \frac{1}{T} \sum_{t = 0}^{T - 1} [\reward{\piopt} - \reward{t}], \nonumber \\ 
&\text{Average constraint violation (\texttt{CV})} := \frac{1}{T} \left[\sum_{t = 0}^{T-1} b - \const{t} \right]_{+}, \nonumber 
\intertext{where $[x]_{+} = \max\{x,0\}$. For this algorithm, we define the \emph{primal regret} and \emph{dual regret} as follows:}
&\cR^{p}(\pi^*, T)  := \sum_{t = 0}^{T-1} \langle \piopt(\cdot|s) - \pit(\cdot|s), \rewardqhat{t}(s,\cdot) + \lambdat \constqhat{t}(s,\cdot)] \rangle_{s\sim \nu_{\rho,\piopt}}, \nonumber \\
&\cR^{d}(\lambda, T)  := \sum_{t = 0}^{T-1} ( \lambdat - \lambda)\, ( \consthat{t} - b)\,.\label{eq:regret}   
\end{align}
Here, $\langle f,g \rangle_{s\sim \nu_{\rho,\piopt}} = \mathbb{E}_{s\sim \nu_{\rho,\piopt}} [f(s)g(s) ]$ 
and $\nu_{\rho,\piopt}$ is the discounted occupation measure induced by following $\piopt$ from $\rho$ normalized so that it becomes a probability measure.
Observe that the above quantities correspond to the regret for online linear optimization algorithms that can independently update the primal and dual variables. Our main result (proved in~\cref{app:proofs}) in this section characterizes the performance of a generic algorithm in terms of its primal and dual regret.  
\begin{restatable}{theorem}{generic}
Assuming that $\norm{\rewardq{t} - \rewardqhat{t}}_{\infty} \leq \tvarepsilon$ and $\norm{\constq{t} - \constqhat{t}}_{\infty} \leq \tvarepsilon$, for a generic algorithm producing a sequence of polices $\{\pi_0, \pi_1, \ldots, \pi_{T-1}\}$ and dual variables $\{\lambda_0, \lambda_1, \ldots, \lambda_{T-1}\}$ such that for all $t$, $\lambdat$ is constrained to lie in the $[0, U]$ where $U > \lambda^*$, \texttt{OG} and \texttt{CV} can be bounded as:
\begin{align*}
\texttt{\text{OG}} & \leq \frac{\cR^{p}(\pi^*, T) + (1 - \gamma) \cR^{d}(0, T)}{(1 - \gamma) T} + \tvarepsilon \, g(U), \\
\texttt{\text{CV}} &\leq \frac{\cR^{p}(\pi^*, T) + (1 - \gamma) \cR^{d}(U, T)}{(U - \lambda^*) (1 - \gamma) T} + \frac{\tvarepsilon \, g(U)}{(U - \lambda^*)} ,
\end{align*}
where $g(U) := \left[\frac{1 + U}{1 - \gamma} + U \right]$.
\label{thm:generic-bound}
\end{restatable}
We note that such a general primal-dual regret decomposition for convex MDPs (including CMDPs) was recently done by~\citet{zahavy2021reward}. However, they handle the tabular setting where the primal variables correspond to state-action occupancy measures, whereas, the above result defines the primal variables to be the policy parameters. More importantly, our result does it require any assumption about the underlying CMDP. In the unconstrained setting, reducing the policy optimization problem to that of online linear optimization has been previously explored in the \textit{Politex} algorithm~\citep{abbasi2019politex}, and we build upon this work. 

In order to bound the average reward optimality gap and the average constraint violation, we need to (i) project the dual variables onto the $[0, U]$ interval and ensure that $U > \lambda^*$, (ii) update the primal and dual variables to control the respective regret in~\cref{eq:regret}, and (iii) control the approximation error $\tvarepsilon$. Next, we use this recipe to design algorithms with provable guarantees.

\section{Instantiating the framework}
\label{sec:framework-instantiation}
In this section, we will instantiate the primal-dual framework by using the above technique -- specifying the value of $U$ in~\cref{sec:u-specification} and describing algorithms that control the primal and dual regret in~\cref{sec:algorithms}.

\subsection{Upper-bound for dual variables}
\label{sec:u-specification}
In~\cref{app:proofs}, we prove the following upper-bound on the optimal dual variable

\begin{restatable}{lemma}{sd}
The objective~\cref{eq:objective} satisfies strong duality, and the optimal dual variables are bounded as $\lambda^* \leq \frac{1}{\zeta (1 - \gamma)}$, where $\zeta := \max_{\pi} \const{\pi} - b > 0$. 
\label{lemma:sd}
\end{restatable}
Unlike~\citet{ding2020natural,ding2021provably} who bound the dual variables in terms of the unknown Slater constant, the upper-bound from~\cref{lemma:sd} can be computed by maximizing the constraint value function as an unconstrained problem. Throughout, we will set $U = \frac{2}{\zeta \, (1- \gamma)}$ that satisfies the requirement $U>\lambda^*$ and projects the dual variables onto $[0, U]$ range (in~\cref{sec:framework}). 

\subsection{Controlling the primal and dual regret}
\label{sec:algorithms}
In this section, we specify two algorithms to update the primal and dual variables, and control the primal and dual regret respectively. In particular, in~\cref{sec:gda}, we will use mirror ascent to update the primal variables, and gradient descent to update the dual variables. Inspired by the literature on online linear optimization~\citep{orabona2016coin}, we will use robust, parameter-free algorithms to update the primal and dual variables in~\cref{sec:cb}.  

\vspace{-2ex}
\subsubsection{Gradient descent ascent}
\label{sec:gda}
At iteration $t \in [T]$, if the primal and dual iterates are $\pit$ and $\lambdat$ respectively, given $\rewardqhat{t}$ and $\constqhat{t}$, the gradient descent ascent (GDA) update \todoc{a bit misleading name; more like mirror ascent, gradient descent} can be written as follows: if $\lagqhat{t}(s,a) = \rewardqhat{t}(s,a) + \lambda_{t} \, \constqhat{t}(s,a)$ and $\consthat{t} = \sum_{s \in \cS} \rho(s) \sum_{a \in \cA} \pit(a | s) \constqhat{t}(s,a)$, then, 
\begin{align}
\pitt(a|s) & = \frac{\pit(a|s) \exp\left(\eta_1 \lagqhat{t}(s,a) \right)}{\sum_{a'}\pit(a'|s)\exp\left(\eta_1 \lagqhat{t}(s,a') \right)} \label{eq:gda-primal} \\
\lambdatt & = \PP_{\left[0, U \right]}[\lambdat - \eta_2 \, (\consthat{t} - b)] \label{eq:gda-dual}.
\end{align}
Here $\PP_{[a,b]}$ is a projection onto the $[a,b]$ interval, and $\eta_1$ and $\eta_2$ are the step-size parameters for the primal and dual updates respectively. In the tabular setting, the resulting algorithm is the same as that analyzed by~\citet{ding2020natural}. 

Analyzing the primal and dual regret for the above updates is fairly standard in online linear optimization. Using results from the paper of~\citet[Theorem 6.8]{orabona2019modern}, by setting $\eta_1 = \sqrt{\frac{2 \log |\cA|}{t}} \, \frac{1 - \gamma}{1 + U}$, $\eta_2 = \frac{U (1 - \gamma)}{\sqrt{t}}$ and $U = \frac{2}{\zeta \, (1- \gamma)}$,
we get
 $\cR^{p}(\pi^*,T) \leq  \frac{1 + U}{1 - \gamma} \sqrt{2 \log |\cA|} \sqrt{T}$ 
 and $\cR^{d}(\lambda,T) \leq \frac{U}{1 - \gamma} \sqrt{T}$. Observe that both the primal and dual regret scale as $O(\sqrt{T})$, and using~\cref{thm:generic-bound}, both the average optimality gap and constraint violation will decrease at an $O(\nicefrac{1}{\sqrt{T}})$ rate. 

We also note that obtaining the above bounds requires setting the two step-sizes ($\eta_1$ and $\eta_2$) to specific values that depend on problem-dependent parameters. \todoc{this does not seem to be the case.. at least the above values are instance independent.}
In~\cref{fig:sensitivity-intro}, we have seen that GDA is quite sensitive to the values of $\eta_1$ and $\eta_2$, even in the simple tabular setting. In order to alleviate this, we use the recent progress in online linear optimization, and propose robust algorithms in the next section. 

\subsubsection{Coin-betting}
\label{sec:cb}
\citet{orabona2016coin} and \citet{orabona2017training} propose \emph{coin-betting} algorithms that reduce the online linear optimization problems in~\cref{eq:regret} to online betting. Unlike adaptive gradient methods like AdaGrad~\citep{duchi2011adaptive} or Adam~\citep{kingma2014adam} that require setting the initial step-size, coin-betting algorithms are completely parameter-free. We will directly instantiate the regret-minimization algorithms from these works. First, we instantiate the algorithm of~\citet{orabona2016coin} for updating the policy (primal variables) in the CMDP setting. In order to do this, we define additional variables $w_{t}$ for each $(s,a)$ pair and iteration $t$. These variables will be computed recursively, and used to compute the policy $\pitt$ at iteration $t$. In particular, for $t \geq 1$,
\begin{align}
& w_{t+1}(s,a) = \frac{\sum_{i=0}^{t} \lagatilde{i}(s,a)}{(t+1) + T/2} \left(1 + \sum_{i=0}^{t} \lagatilde{i}(s,a) \, w_{i}(s,a) \right)  \nonumber \\
& \pitt(a|s) = 
\begin{cases}
    \pi_0(a|s), \quad \text{if } \sum_{a}{\pi_0(a|s) \, [w_{t+1}(s,a)]_{+}} = 0 \\
    \frac{\pi_0(a|s) \, [w_{t+1}(s,a)]_{+}}{\sum_{a'}{\pi_0(a'|s) \, [w_{t+1}(s,a')]_{+}}}, \quad \text{otherwise} 
    \label{eq:cb-primal}
\end{cases}
\end{align}
where, given $\pit$, $\lagatilde{t}(s,a)$ is equal to $$\lagahat{t}(s,a) \, \cI\{w_t(s,a) > 0\} + [\lagahat{t}(s,a)]_{+} \, \cI\{w_t(s,a) \leq 0\}$$ and $\lagahat{t}(s,a) = \frac{1 - \gamma}{1 + U} \, \left[\lagqhat{t}(s,a) - \left \langle \lagqhat{t}(s,\cdot), \pit(\cdot|s) \right \rangle \right]$. $\cI\{\omega\}$ is the indicator function with value $1$ when condition $\omega$ satisfy. For the above calculation, we use the normalized (by $\frac{1 + U}{1 - \gamma}$) action-value functions that are ensured to lie in the $[0,1]$ range. The quantity $\lagahat{t}(s,a)$ can be interpreted as the (normalized) advantage function for policy $\pit$ in the unconstrained MDP with rewards equal to $r(s,a) + \lambdat \, c(s,a)$. Observe that the above update does not have any tunable hyperparameters. 

Similarly, we use the coin-betting algorithm of~\citet{orabona2017training} to update the Lagrange multipliers, instantiating it in the CMDP setting: for $t \geq 1$, if $\sigma(x) := \frac{1}{1 + \exp(-x)}$, then, 
\begin{align}
\lambdatt &= \lambda_0 - \beta_{t} \left[\frac{1}{1 - \gamma} - \sum_{i = 0}^{t} \left(\lambda_i - \lambda_0 \right) \, (\consthat{\pi_i} - b) \right],  \nonumber \\
\beta_{t} & = (1 - \gamma) \, \left( 
2 \sigma \left(\frac{2 \sum_{i = 0}^{t} (\consthat{\pi_i} - b)}{\frac{1}{1 - \gamma} + \sum_{i = 0}^{t} \vert \consthat{\pi_i} - b \vert} - 1 \right) \right) \label{eq:cb-dual} .
\end{align}
Similar to the primal update, the dual update uses normalized (by $\nicefrac{1}{1 - \gamma}$) value functions that lie in the $[-1,1]$ range, and does not have any tunable parameter. Importantly, these updates result in \emph{no-regret} algorithms meaning that both the primal and dual regret scale as $o(T)$. Specifically, for the primal updates in~\cref{eq:cb-primal} and the dual updates in~\cref{eq:cb-dual}, the results of~\citet{orabona2017training} imply that
$$
\cR^{p}(\pi^*,T) \leq \frac{3 (1 + U)}{1 - \gamma} \sqrt{T} \sqrt{1 + \text{KL}(\pi_0 || \pi^*)},
$$ 
$$
\cR^{d}(\lambda,T) \leq \frac{1}{1 - \gamma} + \norm{\lambda-\lambda^0} \sqrt{\left(\frac{1}{(1 - \gamma)^2}+\frac{G_T}{1 - \gamma}\right)
\Gamma_T},$$
where 
$\text{KL}(\pi_0||\pi^*) =\mathbb{E}_{s\sim \nu_{\rho,\piopt}} \text{KL}(\pi_0(\cdot|s)||\piopt(\cdot|s)$,
$\Gamma_T=\log\left(1 + (G_T \, (1-\gamma)+1)^2\norm{\lambda-\lambda^0}^2 \right)$ and $G_T = \sum_{i = 0}^{T} \vert \consthat{\pi_i} - b \vert = O(T)$. Since both regrets scale as $O(\sqrt{T})$ in the worst case, using the coin-betting updates will also result in an $O(\nicefrac{1}{\sqrt{T}})$ decrease in both the average optimality gap and constraint violation. Unlike the updates in~\cref{sec:gda}, the coin-betting updates do not require tuning a hyperparameter.

If we can control the approximation errors, we can use the above algorithms to completely instantiate the primal-dual framework. In~\cref{app:tabular}, we do this for the simpler tabular setting, and consider the linear function approximation setting in the next section. 

\section{Putting everything together}
\label{sec:putting-together}
In this section, we will bound the approximation errors in the linear function approximation setting and instantiate the above framework. 

In order to scale to large state-action spaces, we consider the special case of linear function approximation and assume \emph{global} access to a $d$-dimensional feature map $\Phi : \cS \times \cA \rightarrow \R^d$. Given $\Phi$, we make the following (approximate) realizability assumption on action-value functions~\citep{abbasi2019politex}.  
\begin{assumption}[Linear function approximation]
With global access to the feature map $\Phi$, the action-value functions for each memoryless policy $\pi$ are $\epsb$-close to the span of the state-action features. 
\begin{align*}
\inf_{\theta \in \R^d} \max_{(s,a)} \vert \rewardq{\pi}(s,a) - \langle \theta, \phi(s,a) \rangle \vert & \leq \epsb
\\
\inf_{\theta \in \R^d} \max_{(s,a)} \vert \constq{\pi}(s,a) - \langle \theta, \phi(s,a) \rangle \vert & \leq \epsb
\end{align*}
\label{assum:linear-realizability}
\end{assumption}
\vspace{-2ex}
This setting subsumes the tabular case which can be recovered (with $\epsb = 0$) when $d = |\cS | \, |\cA |$, and the feature-map consisting of one-hot vectors for each state-action pair. Given a good estimate of $\theta_r^\pi := \argmin \left[\max_{(s,a)} \vert \rewardq{\pi}(s,a) - \langle \theta, \phi(s,a) \rangle \vert \right]$, we can easily estimate the action-value functions for every $(s,a)$ pair as $\rewardq{\pi}(s,a) \approx \langle \theta_r^\pi, \phi(s,a) \rangle$. A naive way to estimate $\theta_r^\pi$ is to form a subset $\cC \subseteq \cS \times \cA$ of $(s,a)$ pairs, rollout $m$ independent trajectories using policy $\pi$ and starting from each $(s,a) \in \cC$. The average (across trajectories) cumulative discounted return is an unbiased estimate $q_r(s,a)$ of the action-value function. If $q_r$ is defined to be the $|\cC|$-dimensional vector of estimated action-value functions, and for a fixed set of weights $\omega$ s.t. $\omega(s,a) \geq 0$ and $\sum_{(s,a) \in \cC} \omega(s,a) = 1$, we use the weighted-least squares estimate with $z := (s,a)$,
\begin{align}
\thetahat^\pi_r &= \argmin_{\theta} \sum_{z \in \cC} \omega(z) \left[ \langle \theta, \phi(z) \rangle - q_r(z) \right]^2.
\label{eq:lspg}
\end{align}
For the $(s,a) \in \cC$, the sampling error is $O(\nicefrac{1}{\sqrt{m}})$ by using Hoeffding's inequality. For the $(s,a) \notin \cC$, we can then use the resulting $\thetahat^\pi_r$ to estimate $\rewardqhat{\pi}$ as $\rewardqhat{\pi} = \langle \thetahat^\pi, \phi(s,a) \rangle$. In~\cref{app:proofs}, we prove the following result to bound the extrapolation errors for all $(s,a)$.  
\begin{restatable}{lemma}{faextr}
For policy $\pi$, any distribution $\omega$ and subset $\cC$, if we use $m$ trajectories to estimate the action-value function for each $(s,a) \in \cC$, and solve~\cref{eq:lspg} to compute $\thetahat_r^\pi$, then for any $(s,a) \in (\cS \times \cA)$ pair, the error $\vert \langle \phi(s,a), \thetahat_r^\pi \rangle - \rewardq{\pi} \vert$ can be upper-bounded by   
\begin{align}
\epsb (1+ \indnorm{\phi(s,a)}{G_{\omega}^{\dagger}}) \nonumber + \frac{\indnorm{\phi(s,a)}{G_{\omega}^{\dagger}}}{1 - \gamma} \, \sqrt{\frac{\log(2 |\cC| /\delta)}{2 m}}.  
\end{align}
where, $G_{\omega} = \sum_{(s,a) \in \cC} \omega(s,a) \phi(s,a) \phi(s,a)^{\transpose}$ and $A^\dagger$ is pseudoinverse of $A$.
\label{lemma:lspe-extrapolation}
\end{restatable}
Hence, the extrapolation errors can be upper-bounded by choosing $\cC$ and $\omega$ to control the $\indnorm{\phi(s,a)}{G_{\omega}^{\dagger}}$ term for each $(s,a)$ pair. Moreover, to ensure scalability, we want that size of $\cC$ to be independent of $|\cS||\cA|$. Fortunately, the Kiefer-Wolfowitz theorem~\citep{kiefer1960equivalence} guarantees the existence of a \emph{coreset} $\cC$ s.t. $|\cC| \leq \frac{d (d+1)}{2}$ and distribution $\omega$ that ensure $\sup_{(s,a)} \indnorm{\phi(s,a)}{G_{\omega}^{\dagger}} \leq \sqrt{d}$. If we can find such a $\cC$ and distribution $\omega$, then function approximation error, $\tvarepsilon \leq \epsb (1+\sqrt{d}) + \frac{\sqrt{d}}{1 - \gamma} \sqrt{\frac{\log(2 d (d+1) /\delta)}{2 m}}$. For our theoretical results, we  assume that a coreset $\cC$ and distribution $\omega$ is provided, and in~\cref{app:exp-details}, we describe the G-experimental design procedure to compute it. 

Now that we have control over $\tvarepsilon$, we instantiate the primal-dual framework with coin-betting algorithms. 

\subsection{\Alg/ Algorithm}
\label{sec:lfa-algorithm}
\vspace{-2ex}
\begin{algorithm}[t]
\LinesNumbered
\caption{Coin-Betting Politex}
\label{alg:cbp}
    \textbf{Input}: $\pi_0$ (policy initialization), $\lambda_0$ (dual variable initialization), $m$ (Number of trajectories), $T$ (Number of iterations), Feature map $\Phi$. \vspace{1ex} 
    
    Compute coreset $\cC$ and distribution $\omega$ \vspace{1ex} 
    
    Solve the unconstrained problem $\max_{\pi} \consthat{\pi}$ to estimate $\zeta$ in~\cref{lemma:sd} and set $U = \frac{2}{\zeta (1 - \gamma)}$.  \vspace{1ex}
    
    \For{$t \leftarrow 0$  \KwTo  $T-1$}{
    For every $(s,a) \in \cC$, use $m$ trajectories starting from $(s,a)$ using policy $\pit$ and estimate the action-value functions $q_r(s,a)$ and $q_c(s,a)$. \vspace{1ex}
    
    Compute and store $\thetahat_r^{\pit}$ and $\thetahat_c^{\pit}$ using~\cref{eq:lspg}.  \\

    \For{every $s$ encountered in the trajectory generated by $\pit$, and for every $a$}{\vspace{1ex}
    Compute \label{algline:extrapolation}
    $\rewardqhat{t}(s,a)  = \langle \thetahat^{\pit}_{r}, \phi(s,a) \rangle$; $\constqhat{t}(s,a) = \langle \thetahat^{\pit}_{c}, \phi(s,a) \rangle$ and $\lagqhat{t}(s,a)  = \rewardqhat{t}(s,a) + \lambda_{t} \, \constqhat{t}(s,a)$. \vspace{1ex}
    
    Update $\pitt(a|s)$ using~\cref{eq:cb-primal}. 
    }
    
    Compute $\consthat{\pit}$, update $\lambdatt$ using~\cref{eq:cb-dual}. 
    }
\end{algorithm}
In this section, we use the coin-betting algorithms (\cref{sec:cb}) with linear function approximation to completely specify the Coin-Betting Politex (\Alg/) algorithm (\cref{alg:cbp}). In~\cref{alg:cbp},~Line~$2$ computes the coreset $C$ and distribution $\omega$ offline (see~\cref{app:g-experimental-algo} for details). In order to set $U$, the upper-bound on the dual variables, we need to estimate $\zeta$ and this is achieved by solving the unconstrained problem maximizing $\consthat{\pi}$ in~Line~$3$. While this can be done by any algorithm that can solve MDPs with linear function approximation (for example, NPG~\citep{kakade2001natural} or Politex~\citep{abbasi2019politex}), we will use~\cref{eq:cb-primal} (see~\cref{sec:Experiments}) in this work. After Monte-Carlo sampling $\forall (s,a) \in \cC$ (Line~$5$) and estimating $\thetahat_r^{\pit}$ and $\thetahat_c^{\pit}$ according to~\cref{eq:lspg} (Line~$6$), these vectors are used to calculate $\rewardqhat{\pit}$ and $\constqhat{\pit}$ for states encountered in a trajectory generated by policy $\pit$ (Line~$8$). These action-value functions are then used to update the policy at these states. While this can be achieved by any algorithm controlling the primal regret, \Alg/ uses the parameter-free coin-betting updates (Line~$9$). At the end of iteration $t$, in Line~$11$, the dual variables are updated using the coin-betting algorithm.

In the next section, we bound the average optimality gap and constraint violation for \Alg/.

\vspace{-2ex}
\subsubsection{Theoretical Guarantee}
\label{sec:linear-bound}
We now use~\cref{thm:generic-bound} to bound the average optimality gap and constraint violation for~\cref{alg:cbp}. We note that recent work~\citep{liu2021parameter} uses parameter-free coin-betting algorithms for convex-concave min-max optimization. 
Since the function to be maximized in~\cref{eq:objective-saddle} is non-concave in $\pi$, this work is not directly applicable to our setting.

\begin{restatable}{corollary}{cblfa}
Under~\cref{assum:linear-realizability}, \texttt{OG} and \texttt{CV} of \Alg/ can be bounded as:
\begin{align*}
\texttt{\text{OG}} & \leq \frac{
\left(\frac{3 (1 + U) \, \sqrt{1 + \text{KL}(\pi_0 || \pi^*)}}{1 - \gamma} +  \Psi \right)}{(1 - \gamma) \sqrt{T}}+ \frac{\tvarepsilon (1 + 2 U)}{1 - \gamma}, \\
\texttt{\text{CV}} &\leq  \frac{\zeta \left(\frac{3 (1 + U) \, \sqrt{1 + \text{KL}(\pi_0 || \pi^*)}}{1 - \gamma} +  \Psi \right)}{\sqrt{T}} + \zeta \, \tvarepsilon (1 + 2 U),
\end{align*}
where $U = \frac{2}{\zeta (1 - \gamma)}$, $\tvarepsilon = \epsb (1+\sqrt{d}) + \frac{\sqrt{d}}{1 - \gamma} \sqrt{\frac{\log(2 d (d+1) /\delta)}{2 m}}$ and $\Psi= 4U\sqrt{\log((T+1)U)} + 1$.
\label{cor:cb-lfa}
\end{restatable}  
Since $U = O(\nicefrac{1}{1-\gamma})$, the average optimality gap for \Alg/ is $O \left(\frac{1}{(1 - \gamma)^3 \, \sqrt{T}} + \frac{\tvarepsilon}{(1 - \gamma)^2} \right)$, while the average constraint violation scales as $O \left(\frac{1}{(1 - \gamma)^2 \, \sqrt{T}} + \frac{\tvarepsilon}{1 - \gamma} \right)$. In the function approximation case, ignoring sampling errors,~\citet{ding2020natural} obtain an $O \left(\frac{1}{(1 - \gamma)^3 \sqrt{T}} + \left[\frac{\epsb}{(1 - \gamma)^3} \, \norminf{\frac{d^\pi{^*}}{\rho}} \right]^{1/2} \right)$ average optimality gap, and an $O \left(\frac{1}{(1 - \gamma)^2 T^{1/4}} + \left[\frac{\epsb}{(1 - \gamma)^3} \, \norminf{\frac{d^{\piopt}}{\rho}} \right]^{1/4} \right)$ average constraint violation. Here, $d^{\piopt}$ is the distribution over states induced by the optimal policy, and $\rho$ is the initial state distribution. Compared to~\cref{cor:cb-lfa}, the \texttt{CV} decreases at a slower $O\left(\nicefrac{1}{T^{1/4}}\right)$ rate. Comparing the error terms, the bound for~\citet{ding2020natural} depends on the potentially large (even infinite) $\norminf{\frac{d^{\piopt}}{\rho}}$ factor, while forming the coreset ensures that the errors are well controlled for~\cref{cor:cb-lfa}. Furthermore,~\citet{ding2020natural} require knowledge of the typically unknown Slater constant for the CMDP. 

On the other hand,~\citet{xu2021crpo} use a neural function approximation (with 1 hidden layer) where only the first layer is trained. In order to compare to~\cref{cor:cb-lfa}, we set the width of the second layer to $1$ in~Theorem~2 of \citet{xu2021crpo}, making the function approximation equal to a linear mapping with a ReLU non-linearity. In this setting,~\citet[Theorem 4]{xu2021crpo} prove that both the average optimality gap and constraint violation scale as $O \left( \frac{1}{(1 - \gamma) \sqrt{T}} + \frac{\epsb}{(1 - \gamma)^{2.5}} \, \norminf{\frac{d^{\piopt}}{\rho}} \right)$. 
Observe that although both \texttt{OG} and \texttt{CV} decrease at an $O\left(\nicefrac{1}{\sqrt{T}}\right)$ rate, the error amplification also depends on $\norminf{\frac{d^{\piopt}}{\rho}}$. Furthermore, this result requires setting the hyperparameters according to the typically unknown $\text{KL}(\piopt || \pi_0)$ quantity. These problems make the theoretical results of~\citet{ding2020natural} and \citet{xu2021crpo} potentially vacuous, and the algorithms difficult to use.


\vspace{-1ex}
\section{Experiments}
\label{sec:Experiments}
In this section, we first describe some practical considerations for implementing \red{\Alg/} and compare with baselines \blue{GDA} and \teal{CRPO} on a synthetic tabular environment and the Cartpole environment with linear function approximation. The code can be found at \url{https://github.com/arushijain94/CoinBettingPolitex}.
\vspace{-2ex}
\subsection{Practical considerations}
\label{sec:practical}
\vspace{-2ex}
\paragraph{Checking feasibility and Estimating $\zeta$:}We use the updates in~\cref{eq:cb-primal} to solve the unconstrained problem maximizing $\consthat{\pi}$, and return policy $\pitil$. If $\consthat{\pitil} < b$, we declare the problem infeasible, whereas, if $\consthat{\pitil} > b$, we estimate $\zeta = \consthat{\pitil} - b$.\footnote{If $\consthat{\pitil} = b$, then we return policy $\pitil$ as the optimal feasible policy in the CMDP.} It is important to note that~\cref{lemma:sd} does not require the exact maximization of $\consthat{\pi}$ to upper-bound $\lambda^*$. Any feasible policy for which $\const{\pi} > b$ can be used to estimate $\zeta$ and upper-bound $\lambda^*$, though the tightest upper-bound is obtained for $\max_{\pi} \const{\pi}$ (see the proof of~\cref{lemma:sd} in~\cref{app:proofs}).   

\paragraph{Gradient normalization and practical coin-betting:} Recall that the coin-betting algorithms in~\cref{sec:cb} require normalizing the gradients by $\nicefrac{1 + U}{1 - \gamma}$. Unfortunately, this upper-bound on the gradient norms is quite loose in practice, and directly using the updates~\cref{eq:cb-primal,eq:cb-dual} results in poor empirical performance. Since coin-betting algorithms do not have a step-size that can be scaled to counteract the normalization, this issue needs to be handled differently. In particular, we continue to directly use the updates in~\cref{eq:cb-primal} with the normalization, but use a heuristic, Algorithm 2 of \citealt{orabona2017training}, for updating the dual variable. This heuristic is a way to adaptively normalize the dual gradients (depending on the previously observed values).
For the details, see~\cref{alg:cbp-practical} in \cref{app:practical-cbp-algo}. 
While this heuristic introduces a hyperparameter in the dual updates, our empirical results suggest 
that the resulting coin-betting algorithm is quite robust to the choice of this parameter and so we use this method in our subsequent experiments.
\vspace{-2ex}
\subsection{Tabular Setting}
\label{sec:tabular-exps}
\begin{figure}[h]
		\begin{subfigure}[b]{0.45\linewidth}
			\centering
			\captionsetup{justification=centering}
			\includegraphics[width=0.9\textwidth]{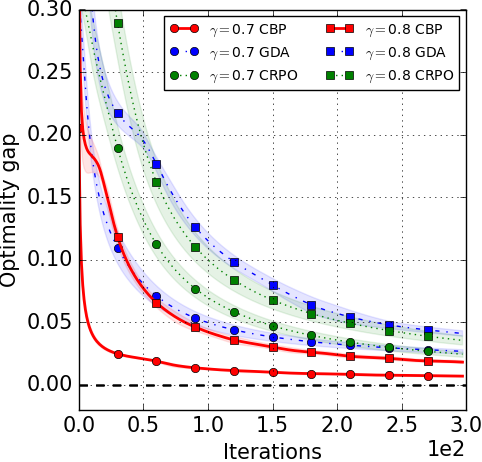}
			\caption[]{\small{OG}}
		\end{subfigure}
		\begin{subfigure}[b]{0.46\linewidth}
			\centering
			\captionsetup{justification=centering}
			\includegraphics[width=0.9\textwidth]{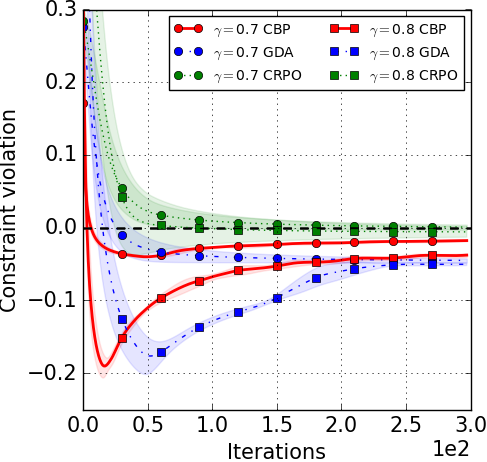}
			\caption[]{\small{CV}}
		\end{subfigure}
		\caption[]
        {\textbf{Environment Misspecification -- Model-based tabular setting with varying $\gamma$:} 
        Assuming access to the true CMDP, we vary discount factor $\gamma = \{0.7, 0.8\}$. We use the hyperparameters for the original CMDP with $\gamma=0.9$. \cAlg/ converges faster with a smaller variance as compared to \cGDA/ and \cCRPO/.}
        \label{fig:CB_GDA_discount_MB}
\end{figure}

We consider a synthetic gridworld environment similar to~\citet[Example 3.5]{sutton2018reinforcement} (see~\cref{app:des-tabular-env} for details) and set the discount factor $\gamma = 0.9$. We first consider a \textbf{model-based setting}, where we have complete knowledge of the CMDP. In~\cref{fig:sensitivity-intro} (in~\cref{sec:Introduction}), we compared the performance of the three algorithms. For each algorithm, the hyperparameter range is described in~\cref{app:experiments} and the \textit{best hyperparameter} corresponds to the least \texttt{OG} while satisfying \texttt{CV} $\in [-0.25,0]$. The key observation is that \textit{CBP is robust to its hyperparameter values}, while GDA and CRPO are sensitive to their hyperparameter values. In~\cref{fig:CB_GDA_best_param_MB} (\cref{app:tab-exp}), we show best performing variants for all methods. In addition, we demonstrate the poor performance of GDA when used with the theoretical step-sizes suggested in~\cref{cor:gda}. Next, we measure the robustness of the algorithms with respect to \textit{environment misspecification} where we vary $\gamma$. In~\cref{fig:CB_GDA_discount_MB}, we observe that \Alg/ has consistently faster convergence with a lower variation in the performance.   

In~\cref{app:tab-exp}, we also consider the \textbf{model-free setting} and approximate the $Q$ functions with sampling. In this case, we demonstrate the consistent superiority of \Alg/ (\cref{fig:CB_GDA_MF_TDSampling}) and its robustness to hyperparameters (\cref{fig:MF_sampling_1_hot_hyperparam_sensitivity}) and environment misspecification (\cref{fig:MF_sampling_1_hot_gamma}).   

\vspace{-2ex}
\subsection{Linear Setting}
\label{sec:linear-exps}
\begin{figure}[h]
		\begin{subfigure}[b]{0.45\linewidth}
			\centering
			\captionsetup{justification=centering}
			\includegraphics[width=0.98\textwidth]{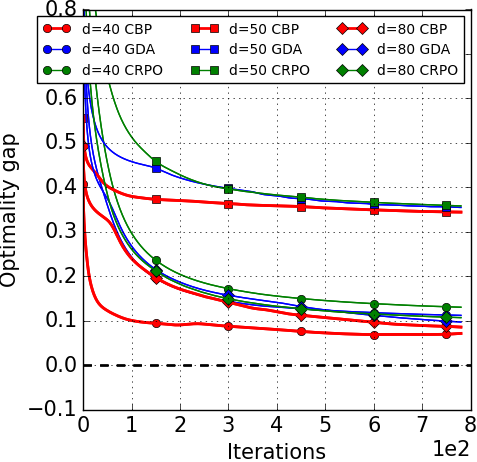}
			\caption[]{\small{OG}}
		\end{subfigure}
		\begin{subfigure}[b]{0.46\linewidth}
			\centering
			\captionsetup{justification=centering}
			\includegraphics[width=0.98\textwidth]{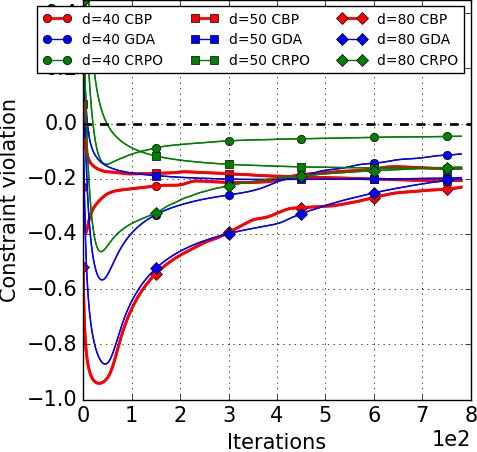}
			\caption[]{\small{CV}}
		\end{subfigure}
		\caption[]
        {\textbf{LFA in gridworld environment:} For varying feature dimension $d$, \texttt{OG} for \cAlg/ consistently converges faster than the baselines \cGDA/ and \cCRPO/.}
        \label{fig:LFA_grid}
\end{figure}

\paragraph{Gridworld environment:}We start with linear function approximation (LFA) on the gridworld environment. We use tile coding~\citep{sutton2018reinforcement} to construct $d$-dimensional feature space (see~\cref{app:lfa-exp} for details). We used LSTDQ~\citep{lagoudakis2003least} \todoc{why?} to estimate $Q$ functions  with $300$ samples for all $(s,a)$ pairs. In~\cref{fig:LFA_grid}, we show the performance of the best hyperparameter (see~\cref{tab:hyper_lfa_sampling_tc} for specific values) for each algorithm. We observe that the \texttt{OG} of \Alg/ converges consistently faster across different feature dimensions. Again, we observe a good hyperparameter robustness of \Alg/ in~\cref{fig:lfa-gridworld-hyperparmeter-sensitivity} (\cref{app:experiments}).~\cref{fig:g-experimental} in~\cref{app:lfa-exp} shows that we can obtain similar performance by using G-experimental design, but at a much lower computational cost. 

\paragraph{Cartpole environment with exploration:}We use the Cartpole environment from the OpenAI gym~\citep{brockman2016openai}, and modify it to include multiple constraints. The agent is rewarded to keep the pole upright, whereas it receives a constraint reward if (1) the cart enters certain areas (x-axis position), or (2) the angle of pole is smaller than a certain threshold (see~\cref{app:lfa-exp} for details). We used tile coding to construct the feature space, and LSTDQ to estimate the $Q$ functions for both reward and constraint reward. 

\begin{figure}[t]
    \centering
		\begin{subfigure}[b]{0.32\linewidth}
			\centering
			\captionsetup{justification=centering}
			\includegraphics[width=0.99\textwidth]{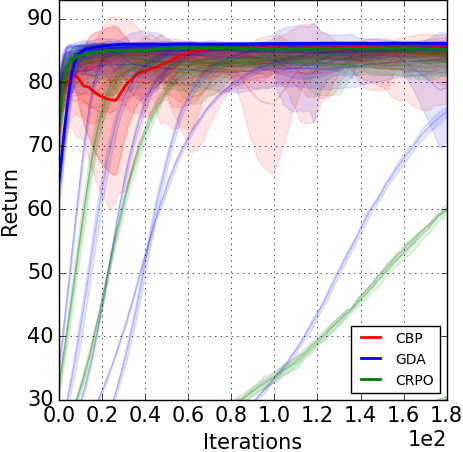}
			\caption[]{\small{Return}}
		\end{subfigure}
		\begin{subfigure}[b]{0.32\linewidth}
			\centering
			\captionsetup{justification=centering}
			\includegraphics[width=0.99\textwidth]{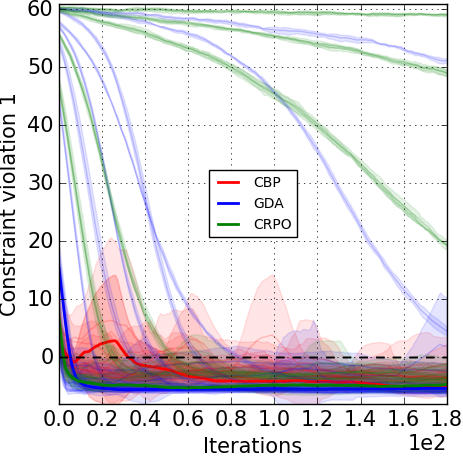}
			\caption[]{\small{CV 1}}
		\end{subfigure}
		\begin{subfigure}[b]{0.32\linewidth}
			\centering
			\captionsetup{justification=centering}
			\includegraphics[width=0.99\textwidth]{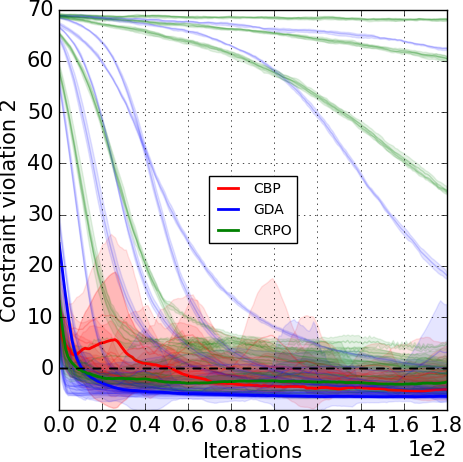}
			\caption[]{\small{CV 2}}
		\end{subfigure}
		\caption[]
        {\textbf{Cartpole environment:} Performance of \cAlg/, \cGDA/ and \cCRPO/ with two constraints (averaged across $5$ runs). The dark lines depict performance with the best hyperparameters. Light lines correspond to performance with other setting of hyperparameters. CBP exhibit robustness to the choice of hyperparameters.}
        \label{fig:cartpole}
\end{figure}

In~\cref{fig:cartpole} we show the cumulative discounted reward and the constraint violation (\texttt{CV 1}, \texttt{CV 2}) for the two constraints as mentioned above. The dark lines correspond to the best hyperparameter that achieves the maximum return, while satisfying  \texttt{CV }$\in [-6,0]$ for both constraints, with the lighter shade-lines correspond to the other hyperparameters. All the three algorithms satisfy the constraints, achieve comparable reward, but \Alg/ has considerably less variance in performance for different values of the hyperparameters. In~\cref{fig:cartpole_entropy_sensitivity} (\cref{app:lfa-exp}), we added entropy regularization ~\citep{geist2019theory,haarnoja2018soft} and observed a similar robustness for CBP.



\vspace{-1ex}
\section{Conclusion}
\label{sec:Conclusion}
\vspace{-1ex}
In this paper, we proposed a general primal-dual framework to solve CMDPs with function approximation. We instantiated this framework using coin-betting algorithms from online linear optimization, and proposed the \Alg/ algorithm. We proved that \Alg/ is not only theoretically sound and has good empirical performance, but also robust to hyper-parameter tuning and environment misspecification. 
In the future, we hope to use the recent advances in online linear optimization to design ``painless'' parameter-free policy optimization algorithms. We believe that this ambition is important for reproducibility in RL, and hope that our work will encourage future research in this area. 
\todoc{
Private summary:
We show the power of designing RL methods
using the reduction that relates policy search to online linear optimization.
We show this approach also works for CMDPs.
With this, 
for the first time we show $1/\sqrt{T}$ rates for the optimality gap and constraint violation.
This also refutes the conjecture of Xu et al.
that primal-dual methods are inferior to primal only methods in this setting in the sense 
that primal-dual methods now have the same rate as the primal only methods.
Another contribution we make is to investigate whether the so-called coin-betting algorithms from online linear optimization, which aim to be parameter-free, lead to more robust performance in our 
RL context. For this, we instantiate these algorithms, and empirically investigate them
and find that they are indeed less sensitive to their remaining hyperparameters than 
algorithms that are derived from simpler approaches to linear optimization.
The price paid is the increase in running time. (is this even true? I guess Politex is also quadratic in $T$, unless one does something more clever.)
And we should have explained that the dependence on $\zeta$ is expected and nuances such as first solving for the constraint reward maximizing value.

What Xu et al wrote:
``the primal-dual approach can be sensitive to the
initialization of Lagrange multipliers and the learning rate,
and can thus incur extensive cost in hyperparameter tuning''.
We partially confirm this and then show that coin betting can help with this.
}
\begin{acknowledgements}
We would like to thank Tor Lattimore for feedback on the paper.

Csaba Szepesv\'ari gratefully acknowledges the funding from Natural Sciences and Engineering Research Council (NSERC) of Canada and ``Design.R AI-assisted CPS Design'' (DARPA)  project. Doina Precup and Csaba Szepesv\'ari both gratefully acknowledge funding from Canada CIFAR AI Chairs Program for Mila and AMII respectively.
\end{acknowledgements}

\bibliography{ref}

\clearpage
\onecolumn
\appendix
\newcommand{\appendixTitle}{%
\vbox{
    \centering
	\hrule height 4pt
	\vskip 0.2in
	{\LARGE \bf Supplementary material}
	\vskip 0.2in
	\hrule height 1pt 
}}

\appendixTitle

\section*{Organization of the Appendix}
\begin{itemize}

  \item[\ref{app:tabular}] \nameref{app:tabular}
    
  \item[\ref{app:proofs}] \nameref{app:proofs}
 
  \item[\ref{app:exp-details}] \nameref{app:exp-details}
  
  \item[\ref{app:experiments}] \nameref{app:experiments}
 
\end{itemize}
\section{Theoretical Guarantees in the Tabular Setting}
\label{app:tabular}
In the tabular setting, we use $m$ independent trajectories for \emph{each} $(s,a)$ pair. By Hoeffding's inequality and union bound across all states and actions, the sampling error can be bounded by $\frac{1}{1 - \gamma} \sqrt{\frac{\log(2 S A/\delta)}{2 m}}$ (similar to the proof of~\cref{lemma:lspe-extrapolation}). Since all action-value functions can be represented in the tabular setting, the bias error term $\epsb = 0$, and hence $\tvarepsilon= \frac{1}{1 - \gamma} \sqrt{\frac{\log(2 S A/\delta)}{2 m}}$. Compared to the linear function approximation setting in~\cref{sec:putting-together} that has a computational complexity proportional to $O(d^2)$, the computational cost in the tabular setting is $O(SA)$. However, the approximation error is smaller than that in~\cref{lemma:lspe-extrapolation}. 

Now that we have bounded the approximation errors in the tabular setting, we instantiate~\cref{thm:generic-bound} for GDA in~\cref{sec:gda}. Plugging in the value of $U$, the primal and dual regret from~\cref{eq:regret} and $\tvarepsilon$, we obtain the following corollary. 
\begin{thmbox}
\begin{restatable}{corollary}{gda}
For the gradient descent ascent updates in~\cref{eq:gda-primal,eq:gda-dual} with the specified step-sizes, $U = \frac{2}{\zeta \, (1 - \gamma)}$, using $m$ trajectories, the average optimality gap (\texttt{OG}) and constraint violation (\texttt{CV}) can be bounded as:
\begin{align*}
\texttt{\text{OG}} & \leq \frac{
\left(\frac{(1 + U) \, \sqrt{2 \log |A|}}{1 - \gamma} + U \right)}{(1 - \gamma) \sqrt{T}} + \frac{\epss (1 + 2 U)}{1 - \gamma}, \\
\texttt{\text{CV}} &\leq  \frac{\zeta \left(\frac{(1 + U) \, \sqrt{2 \log |A|}}{1 - \gamma} + U \right)}{\sqrt{T}} + \epss (1 + 2 U), 
\end{align*}
where $\epss = \frac{1}{1 - \gamma} \sqrt{\frac{\log(2 S A/\delta)}{2 m}}$. 
\label{cor:gda}
\end{restatable}
\end{thmbox}
\begin{proof}
To get the result we replace the regrets for primal and dual of GDA~\citep[Theorem 6.8]{orabona2019modern} in ~\cref{thm:generic-bound} and get the required results. Specifically we set 
\[
\cR^{p}(\pi^*,T) \leq  \frac{1 + U}{1 - \gamma} \sqrt{2 \log |A|} \sqrt{T},
\]
and
\[
\cR^{d}(0,T), \cR^{d}(U,T)  \leq \frac{U}{1 - \gamma} \sqrt{T}.
\]

\end{proof}
Hence, the average optimality gap for GDA is $O \left(\frac{1}{(1 - \gamma)^3 \, \sqrt{T}} + \frac{\epss}{(1 - \gamma)^2} \right)$, while the average constraint violation scales as $O \left(\frac{1}{(1 - \gamma)^2 \, \sqrt{T}} + \frac{\epss}{1 - \gamma} \right)$. Compared to the tabular result in~\citet{ding2020natural}, the above bound on the optimality gap is worse by a factor of $O(\nicefrac{1}{1 - \gamma})$ and matches their bound on the constraint violation. On the other hand, in the tabular setting without sampling error (when $\epss = 0$),~\citet[Theorem 3]{xu2021crpo} obtain an $O \left(\frac{1}{(1 - \gamma)^{1.5} \, \sqrt{T}} \right)$ bound on both the optimality gap and constraint violation. However, in order to set this bound, they require the knowledge of $\text{KL}(\piopt || \pi_0 )$ to set the algorithm hyper-parameters. This information is not available, making it difficult to implement their algorithm. 

Now, we instantiate~\cref{thm:generic-bound} for the coin-betting algorithms in~\cref{sec:cb}. Plugging in the value of $U$, the primal and dual regret and $\tvarepsilon$, we obtain the following corollary. 
\begin{thmbox}
\begin{restatable}{corollary}{cb}
Using the primal updates in~\cref{eq:cb-primal}, and the dual updates in~\cref{eq:cb-dual}, with $U = \frac{2}{\zeta \, (1 - \gamma)}$, using $m$ trajectories, the average optimality gap (\texttt{OG}) and constraint violation (\texttt{CV}) can be bounded as:
\begin{align*}
\texttt{\text{OG}} & \leq \frac{
\left(\frac{3 (1 + U) \, \sqrt{1 + \text{KL}(\pi_0 || \pi^*)}}{1 - \gamma} + \Psi \right)}{(1 - \gamma) \sqrt{T}} + \frac{\epss (1 + 2 U)}{1 - \gamma}, \\
\texttt{\text{CV}} &\leq  \frac{\zeta \left(\frac{3 (1 + U) \, \sqrt{1 + \text{KL}(\pi_0 || \pi^*)}}{1 - \gamma} + \Psi \right)}{\sqrt{T}} + \zeta \, \epss (1 + 2 U), 
\end{align*}
where $\epss = \frac{1}{1 - \gamma} \sqrt{\frac{\log(2 S A/\delta)}{2 m}}$ and $ \Psi= 4U\sqrt{\log((T+1)U)} + 1$.
\label{cor:cb}
\end{restatable}
\end{thmbox}
\begin{proof}
To get the result we replace the regrets for primal and dual of CB in ~\cref{thm:generic-bound} and get the required results. Specifically from~\citep[Corollary 6]{orabona2016coin} and~\citep[Theorem 8]{orabona2017training}, we get the upper-bound for primal regret and the dual regret:    
\[
\cR^{p}(\pi^*,T) \leq \frac{3 (1 + U)}{1 - \gamma} \sqrt{T} \sqrt{1 + \text{KL}(\pi_0 || \pi^*)},
\]
and
\[
\cR^{d}(\lambda,T) \leq \frac{1}{1 - \gamma} + \norm{\lambda-\lambda^0} \sqrt{\left(\frac{1}{(1 - \gamma)^2}+\frac{G_T}{1 - \gamma} \right)
\Gamma_T}
\] 
where $\Gamma_T=\log\left(1 + (G_T \, (1-\gamma)+1)^2\norm{\lambda-\lambda^0}^2 \right)$ and $G_T = \sum_{i = 0}^{T} \vert \consthat{\pi_i} - b \vert$. Since $\vert \consthat{\pi_i} - b \vert \leq \frac{1}{1-\gamma}$ we have $G_T \leq T/1-\gamma$ and $\norm{\lambda - \lambda^0} \leq 2U$ for all $\lambda$. Using these upperbound and replace in $\cR^{d}(\lambda,T)$ we get: 
\[
\cR^{d}(\lambda,T) \leq \frac{4U\sqrt{(T+1)\log((T+1)U)} + 1}{1-\gamma}
\]
\end{proof}
\clearpage
\section{Main Proofs}
\label{app:proofs}

The following well known result will be useful:
\begin{thmbox}
\begin{lemma}[Value difference lemma]
For any value function $\vals{\pi}$ (reward or cost), and any two memoryless policies $\pi$ and $\pip$,
\begin{align*}
\vals{\pip} - \vals{\pi} & = \left(I - \gamma P_{\pip} \right)^{-1} \left[T_{\pip} \vals{\pi} - \vals{\pi} \right] 
\end{align*}
where $T_{\pip} \vals{\pi} = [r_{\pip} + \gamma P_{\pip} \vals{\pi}]$ is the Bellman operator for policy $\pip$. 
\label{lemma:val-diff}
\end{lemma}
\end{thmbox}
\begin{proof}
As is well known, $\vals{\pip} = \left(I - \gamma P_{\pip} \right)^{-1} r_{\pip}$. Hence,
\begin{align*}
\vals{\pip} - \vals{\pi} & =  \left(I - \gamma P_{\pip} \right)^{-1} r_{\pip} - \vals{\pi}  \\
& =  \left(I - \gamma P_{\pip} \right)^{-1} \left(r_{\pip} - \left(I - \gamma P_{\pip} \right) \vals{\pi} \right) \\
& = \left(I - \gamma P_{\pip} \right)^{-1} \left(r_{\pip} + \gamma P_{\pip} \vals{\pi} - \vals{\pi} \right) \\
& = \left(I - \gamma P_{\pip} \right)^{-1} \left[T_{\pip} \vals{\pi} - \vals{\pi} \right] 
\end{align*}
\end{proof}

Let us now turn to the proof of \cref{thm:generic-bound}:

\myquote{\begin{thmbox}
\generic*
\end{thmbox} 
}
\begin{proof}
We will begin with bounding the value differences in the Lagrangian using~\cref{lemma:val-diff}. Let $T_{\piopt}^{r}$ and $T_{\piopt}^{c}$ be the Bellman operators of the optimal policy for the reward and cost respectively. Then, 
\begin{align*}
[\rewards{\piopt} - \rewards{\pit}] + \lambdat \,[\consts{\piopt} - \consts{\pit}] 
&= \left(I - \gamma P_{\piopt} \right)^{-1} \big[ \left[T^{r}_{\piopt} \rewards{\pit} - \rewards{\pit} \right] + \lambdat \left[T^{c}_{\piopt} \consts{\pit} - \consts{\pit} \right] \big] 
\end{align*}
Let $M_{\pi}$ be the state-action operator applied $Q$ functions such that $M_{\pi}(Q)(s) = \sum_{a} \pi(a|s) Q(s,a)$. Observe that $T^{r}_{\piopt} \rewards{\pit} = M_{\piopt} \rewardq{\pit}$ and $\rewards{\pit} = M_{\pit} \rewardq{\pit}$. The expressions for the constraint rewards are analogous. Rewriting the above expression, 
\begin{align*}
[\rewards{\piopt} - \rewards{\pit}] + \lambdat \,[\consts{\piopt} - \consts{\pit}] &= \left(I - \gamma P_{\piopt} \right)^{-1} \bigg[ \left[M_{\piopt} \rewardq{\pit} - M_{\pit} \rewardq{\pit} \right] + \lambdat \left[M_{\piopt} \constq{\pit} - M_{\pit} \constq{\pit} \right] \bigg] \\
&= \left(I - \gamma P_{\piopt} \right)^{-1} \bigg[ [M_{\piopt} - M_{\pit}] \, [\rewardq{\pit} + \lambda_t \constq{\pit}] \bigg] \\
& = \left(I - \gamma P_{\piopt} \right)^{-1} \bigg[ [M_{\piopt} - M_{\pit}] \, [\rewardqhat{\pit} + \lambda_t \constqhat{\pit}] \bigg] \\ 
& + \underbrace{\left(I - \gamma P_{\piopt} \right)^{-1} \bigg[ [M_{\piopt} - M_{\pit}] \, [\rewardq{\pit} - \rewardqhat{\pit} + \lambda_t (\constq{\pit} - \constqhat{\pit})] \bigg]}_{\text{Error}} \\
\end{align*}
Let us first bound the maximum norm of the ``Error'' term, 
\begin{align*}
\norm{\text{Error}}_\infty 
& =
\norm{
 \left(I - \gamma P_{\piopt} \right)^{-1} \bigg[ [M_{\piopt} - M_{\pit}] \, [\rewardq{\pit} - \rewardqhat{\pit} + \lambda_t (\constq{\pit} - \constqhat{\pit})] \bigg]}_\infty\\
 &\leq \frac{1}{1 - \gamma} \norm{[\rewardq{\pit} - \rewardqhat{\pit} + \lambda_t (\constq{\pit} - \constqhat{\pit})]}_{\infty} \\
& \leq \frac{1}{1 - \gamma} \norm{\rewardq{\pit} - \rewardqhat{\pit}}_{\infty} + \lambda_t \norm{\constq{\pit} - \constqhat{\pit}]}_{\infty} \\
\intertext{By assumption, $\norm{\rewardq{\pit} - \rewardqhat{\pit}}_{\infty}$, $\norm{\rewardq{\pit} - \rewardqhat{\pit}}_{\infty}\le \epsilon$.}
\implies \norm{\text{Error}}_\infty \leq \frac{\epsilon}{1 - \gamma} (1 +  \lambdat) &
\intertext{Since the dual variables are projected onto the $[0,U]$ interval, $\lambda_t \leq U$, implying that}
\norm{\text{Error} }_\infty \leq \frac{\epsilon}{1 - \gamma} \left(1 + U \right) & 
\end{align*}
Substituting in this bound on the error, using the convention that left-multiplication by a measure means integration with respect to it,
\begin{align*}
[\reward{\piopt} - \reward{\pit}] + \lambdat \,[\const{\piopt} - \const{\pit}] 
& \leq 
\rho \left(I - \gamma P_{\piopt} \right)^{-1} \bigg[ [M_{\piopt} - M_{\pit}] \, [\rewardqhat{\pit} + \lambda_t \constqhat{\pit}] \bigg] + \frac{\epsilon}{1 - \gamma} \left(1 + U \right) \\
& \leq \frac{1}{1 - \gamma} \nu_{\rho,\pi^*} \bigg[ [M_{\piopt} - M_{\pit}] \, [\rewardqhat{\pit} + \lambda_t \constqhat{\pit}] \bigg] + \frac{\epsilon}{1 - \gamma} \left(1 + U \right)\,, \\
\intertext{
where $\nu_{\rho,\pi^*} = (1-\gamma)\rho \left(I-\gamma P_{\piopt} \right)^{-1}$ is the
discounted probability measure over the states obtained when starting from $\rho$ and following $\piopt$.
Summing from $t = 0$ to $T-1$ and dividing by $T$.}
\frac{1}{T} \nu_{\rho,\pi^*}  \sum_{t = 0}^{T-1} \left[ [\reward{\piopt} - \reward{\pit}] + \lambdat \,[\const{\piopt} - \const{\pit}] \right] &\leq 
\frac{\nu_{\rho,\pi^*}}{(1 - \gamma) T} \sum_{t = 0}^{T-1} \left[ [\cM_{\piopt} - \cM_{\pit}] [\rewardqhat{\pit} + \lambdat \constqhat{\pit}] \right] + \frac{\epsilon}{1 - \gamma} \left(1 + U \right) \\
\end{align*}
Now, observe that 
\begin{align*}
\nu_{\rho,\pi^*} \sum_{t = 0}^{T-1} \left[ 
[\cM_{\piopt} - \cM_{\pit}] [\rewardqhat{\pit} + \lambdat \constqhat{\pit}] \right] &= \sum_{t = 0}^{T-1} \langle \piopt(\cdot|s) - \pit(\cdot|s), \rewardqhats{\pit} + \lambdat \constqhats{\pit}] \rangle_{s\sim \nu_{\rho,\pi^*} } = \cR^{p}(\pi^*, T)     
\end{align*}
Putting everything together, 
\begin{align}
\frac{1}{T} \sum_{t = 0}^{T - 1} [\reward{\piopt} - \reward{\pit}] + \frac{1}{T} \sum_{t = 0}^{T - 1} \lambdat \, [\const{\piopt} - \const{\pit}] & \leq \frac{\cR^{p}(\pi^*, T)}{(1 - \gamma) T} + \frac{\epsilon}{1 - \gamma} \left(1 + U \right).
\label{eq:lagrangian-bound}
\end{align}
The above result bounds the sub-optimality in the Lagrangian. Next, we will see how this result implies a bound on the sub-optimality in the objective and the constraint violation. To bound the reward sub-optimality, we will upper bound the negative of the second term on the left-hand side in the above equation, i.e., we upper bound $\frac{1}{T} \sum_{t = 0}^{T - 1} \lambdat \, [\const{\pit} - \const{\piopt}]$. We have,
\begin{align}
\frac{1}{T} \sum_{t = 0}^{T - 1} \lambdat \, [\const{\pit} - \const{\piopt}] & \leq \frac{1}{T} \sum_{t = 0}^{T - 1} \lambdat \, [\const{\pit} - b] & \tag{since $\const{\piopt} \geq b$} \\
& = \frac{1}{T} \sum_{t = 0}^{T - 1} \lambdat \, [\const{\pit} - \consthat{\pit}] + \frac{1}{T} \sum_{t = 0}^{T - 1}  \lambdat [\consthat{\pit} - b] \nonumber \\
& = \frac{1}{T} \sum_{t = 0}^{T - 1} \lambdat \, [\const{\pit} - \consthat{\pit}] + \frac{\cR^{d}(0, T)}{T}  \nonumber \\
& \leq U \epsilon + \frac{\cR^{d}(0, T)}{T}\,. & \label{eq:cost-bound}
\end{align}
Using~\cref{eq:lagrangian-bound,eq:cost-bound},
\begin{align}
\texttt{OG} = \frac{1}{T} \sum_{t = 0}^{T - 1} [\reward{\piopt} - \reward{\pit}] & \leq
\frac{\cR^{p}(\pi^*, T) + (1 - \gamma) \cR^{d}(0, T)}{(1 - \gamma) T} + \frac{\epsilon}{1 - \gamma} \left(1 + U \right) + U \epsilon
\label{eq:reward-result}
\end{align}
This proves the first part of the theorem. We now bound the constraint violation. For an arbitrary $\lambda$, 
\begin{align}
\frac{1}{T} \sum_{t = 0}^{T - 1} \left[(\lambdat - \lambda) (\const{\pit} - b) \right]  & = \frac{1}{T} \sum_{t = 0}^{T - 1} \left[(\lambdat - \lambda) (\const{\pit} - \consthat{\pit}) \right] + \frac{1}{T} \sum_{t = 0}^{T - 1} \left[(\lambdat - \lambda) (\consthat{\pit} - b) \right] \nonumber \\
& = \frac{1}{T} \sum_{t = 0}^{T - 1} \left[(\lambdat - \lambda) (\const{\pit} - \consthat{\pit}) \right] + \frac{\cR^{d}(\lambda, T)}{T}\,, \nonumber \\
\intertext{implying}
\frac{1}{T} \sum_{t = 0}^{T - 1} \left[(\lambdat - \lambda) (\const{\pit} - b) \right] & \leq U \epsilon + \frac{\cR^{d}(\lambda, T)}{T}\,. & \label{eq:cost-bound-2}
\end{align}
Adding~\cref{eq:cost-bound-2} and~\cref{eq:lagrangian-bound} and reordering the terms gives
\begin{align*}
& \frac{1}{T}  \sum_{t = 0}^{T-1} (\reward{\piopt} - \reward{\pit}) +  \frac{\lambda}{T} \sum_{t = 0}^{T-1} (b - \const{\pit}) \\ 
& \leq \frac{1}{T} \sum_{t = 0}^{T-1} \underbrace{\lambdat (b - \const{\piopt})}_{\leq 0 \text{ since $\const{\piopt} \geq b$.}} + \underbrace{\frac{\cR^{p}(\pi^*, T) + (1 - \gamma) \cR^{d}(\lambda, T)}{(1 - \gamma) T} + \frac{\epsilon}{1 - \gamma} \left(1 + U \right) + U \epsilon}_{h(\lambda)} \\
\implies & \frac{1}{T} \sum_{t = 0}^{T-1} (\reward{\piopt} - \reward{\pit}) +  \frac{\lambda}{T} \sum_{t = 0}^{T-1} (b - \const{\pit}) \leq h(\lambda)
\end{align*}
We consider two cases: (i) if $\sum_{t = 0}^{T-1} (b - \const{\pit}) \geq 0$, we set $\lambda = U$, else, if (ii) $\sum_{t = 0}^{T-1} (b - \const{\pit}) < 0$, we set $\lambda = 0$. Using these choices, and since $\cR^{d}(\lambda,T)$ is linearly increasing in $\lambda$,
\begin{align*}
\frac{1}{T}  \sum_{t = 0}^{T-1} (\reward{\piopt} - \reward{\pit}) + \frac{ U}{T}
\left[ \sum_{t = 0}^{T-1} (b - \const{\pit})\right]_{+} & \leq  h(U) 
\end{align*}
Now take the policy $\pip$ such that $\reward{\piopt} - \reward{\pip} = \frac{1}{T}  \sum_{t = 0}^{T-1} (\reward{\piopt} - \reward{\pit})$ and $\const{\piopt} - \const{\pip} = \frac{1}{T}  \sum_{t = 0}^{T-1} (b - \const{\pit})$. Then,
\begin{align*}
[\reward{\piopt} - \reward{\pip}] + U\left[b - \const{\pip}\right]_{+} & \leq h(U)\,.
\end{align*}
Using~\cref{lemma:lag-constraint} with $C = U > \lambda^*$ and $\beta = h(U)$, we get
\begin{align*}
\texttt{CV} &= \frac{1}{T} \left[\sum_{t = 0}^{T-1} b - \const{\pit}\right]_{+} 
=
\left[b - \const{\pip}\right]_{+} \\
& \leq \frac{h(U)}{U - \lambda^*} 
= \frac{\cR^{p}(\pi^*, T) + (1 - \gamma) \cR^{d}(U, T)}{(U - \lambda^*) (1 - \gamma) T} + \frac{1}{(U - \lambda^*) } \left[\frac{\epsilon}{(1 - \gamma)} \left(1 + U \right) + U \epsilon \right] \,,
\end{align*}
which completes the proof subject to proving \cref{lemma:lag-constraint}.
\end{proof}


\begin{thmbox}
\begin{lemma}[Constraint violation bound]
For any $C > \lambda^*$  and any $\pitil$ s.t. $\reward{\piopt} - \reward{\pitil} + C [b - \const{\pitil}]_{+} \leq \beta$, we have $[b - \const{\pitil}]_{+} \leq \frac{\beta}{C - \lambda^*}$. 
\label{lemma:lag-constraint}
\end{lemma}
\end{thmbox}
\begin{proof}
Define $\nu(\tau) = \max_{\pi} \{\reward{\pi} \mid \const{\pi} \geq b + \tau \}$ and note that by definition, $\nu(0) = \reward{\piopt}$ and that $\nu$ is a decreasing function for its argument.

Let $\lag{\pi}{\lambda} = \reward{\pi}+\lambda(\const{\pi}-b)$. Then,
for any policy $\pi$ s.t. $\const{\pi} \geq b + \tau$, we have
\begin{align}
\lag{\pi}{\lambda^*} & \leq \max_{\pi'} \lag{\pi'}{\lambda^*} \nonumber \\
&= \reward{\piopt} &\tag{by strong duality} \\ 
& = \nu(0) & \tag{from above relation} \\
\implies \nu(0) - \tau \lambda^* & \geq \lag{\pi}{\lambda^*} - \tau \lambda^* = \reward{\pi} + \lambda^* \underbrace{(\const{\pi} - b - \tau)}_{\text{Positive}} \nonumber \\
\implies \nu(0) - \tau \lambda^* & \geq \max_{\pi} \{\reward{\pi} \mid \const{\pi} \geq b + \tau \} = \nu(\tau) \,.\nonumber \\
\implies \tau \lambda^* \leq \nu(0) - \nu(\tau)\,. \label{eq:inter-1}
\end{align}

Now we choose $\tautil = -(b - \const{\pitil})_{+}$.
\begin{align*}
(C - \lambda^*) |\tautil| &= \lambda^* \tautil + C |\tautil| & \tag{since $\tautil \leq 0$} \\
& \leq \nu(0) - \nu(\tautil) + C |\tautil| & \tag{\cref{eq:inter-1}} \\
& = \reward{\piopt} - \reward{\pitil} + C |\tautil| + \reward{\pitil} - \nu(\tautil) & \tag{definition of $\nu(0)$} \\
& = \reward{\piopt} - \reward{\pitil} + C (b - \const{\pitil})_{+} + \reward{\pitil} - \nu(\tautil) \\
& \leq \beta + \reward{\pitil} - \nu(\tautil)\,.
\intertext{Now let us bound $\nu(\tautil)$:} 
\nu(\tautil) & = \max_{\pi} \{\reward{\pi} \mid \const{\pi} \geq b - (b - \const{\pitil})_{+} \}  \\
& \geq \max_{\pi} \{\reward{\pi} \mid \const{\pi} \geq \const{\pitil} \} & \tag{tightening the constraint} \\
\nu(\tautil) & \geq \reward{\pitil} 
\implies (C - \lambda^*) |\tautil| 
 \leq \beta \implies (b - \const{\pitil})_{+} \leq \frac{\beta}{C - \lambda^*} 
\end{align*}
\end{proof}

\subsection{Proof of Lemma~\ref{lemma:sd}}

\myquote{\begin{thmbox}
\sd*
\end{thmbox} 
}
\begin{proof}
Starting from the Lagrangian form in~\cref{eq:objective-saddle},
\begin{align*}
\reward{*} & := \max_{\pi} \min_{\lambda \geq 0} \reward{\pi} + \lambda [\const{\pi} - b] \\
\intertext{Using the linear programming formulation of CMDPs in terms of the state-occupancy measures $\mu$, we know that both the objective and the constraint are linear functions of $\mu$, and strong duality holds w.r.t $\mu$. Since $\mu$ and $\pi$ have a one-one mapping, we can switch the min and the max~\citep{paternain2019constrained}, implying,} 
\reward{*}& = \min_{\lambda \geq 0} \max_{\pi} \reward{\pi} + \lambda [\const{\pi} - b] \\
& = \max_{\pi} \reward{\pi} + \lambda^* [\const{\pi} - b]\,. \\
\intertext{Define $\pitil := \argmax_{\pi} \const{\pi}$. Then,}
\reward{*}& \geq \reward{\pitil} + \lambda^* [\const{\pitil} -b] \\
\implies \lambda^* & \leq \frac{\reward{*} - \reward{\pitil}}{[\const{\pitil} -b]} \leq \frac{1}{(1 - \gamma) \zeta}\,.
\end{align*}
\end{proof}

\subsection{Proofs for Section~\ref{sec:putting-together}}

\myquote{\begin{thmbox}
\faextr*
\end{thmbox} 
}
\begin{proof}
By solving $\thetahat^\pi_r = \argmin_{\theta} \sum_{z \in \cC} \omega(z) \left[ \langle \theta, \phi(z) \rangle - q_r(z) \right]^2$ (~\cref{eq:lspg}), we get that
\[
    \theta_r^\pi = G_{\omega}^{\dagger} \sum_{(s,a) \in \cC} \omega(s,a) \phi(s,a) \, q_r^\pi(s,a)
\]
and lets denote $z=(s,a)$, $\phi(z)=\phi(s,a)$ and $\epsilon(z)=q_r(z)-\rewardq{\pi}(z)+\rewardq{\pi}(z)- \langle \phi(z), \theta_r^* \rangle$ and $\theta^*_r := \argmin_{\theta} \max_{(s,a)} \norm{Q_r^\pi(s,a) - \langle \theta, \phi(s,a) \rangle}$ is the optimal parameter for the given policy $\pi$. 

Therefore we can write $q_r(z)= \epsilon(z)+\langle \phi(z), \theta_r^* \rangle$ 
\begin{align*}
\vert \langle \phi(z), \theta_r^\pi \rangle - \rewardq{\pi} \vert & =\vert \langle \phi(z), \theta_r^\pi \rangle-\langle \phi(z), \theta_r^* \rangle+\langle \phi(z), \theta_r^* \rangle - \rewardq{\pi} \vert \\
& \leq \vert \langle \phi(z), \theta_r^\pi \rangle - \langle \phi(z), \theta_r^* \rangle \vert + \epsilon_b \quad [\epsilon_b \text{ from } \cref{assum:linear-realizability}]
\end{align*}
Now we need to bound the first term of above inequality. Based on the definition of $\epsilon(z)$, we can write $\theta_r^\pi = G_{\omega}^{\dagger} \sum_{z' \in \cC} (\langle \phi(z'), \theta_r^* \rangle + \epsilon(z')) \, \omega(z') \phi(z')$. Using this equality we can get easily that: 
\begin{align*}
\vert \langle \phi(z), \theta_r^\pi\rangle - \langle \phi(z), \theta_r^* \rangle \vert& =\vert  \sum_{z' \in \cC} \epsilon(z') \omega(z') \phi(z)^T G_{\omega}^{\dagger} \phi(z')\vert\\ 
& \leq \sum_{z' \in \cC} \vert \epsilon(z') \vert \vert \omega(z') \phi(z)^T G_{\omega}^{\dagger} \phi(z')\vert  \\
&\leq \vert \max_{z' \in \cC} \epsilon(z')\vert \sum_{z' \in \cC} \vert \omega(z') \phi(z)^T G_{\omega}^{\dagger} \phi(z') \vert \\
\end{align*}
To bound the sum term we can 
\begin{align*}
    \left (\sum_{z' \in \cC} \omega(z') \vert \phi(z)^T G_{\omega}^{\dagger} \phi(z') \vert\right)^2 & \leq \sum_{z' \in \cC} \omega(z') \left ( \vert\phi(z)^T G_{\omega}^{\dagger} \phi(z') \vert\right)^2 & \tag{Jensen's inequality} \\
    &= \phi(z)^T G_{\omega}^{\dagger} \left(  \sum_{z' \in \cC}  \left[\omega(z') \phi(z') \phi(z')^T\right]  \right) G_{\omega}^{\dagger} \phi(z)\\
    & = \|\phi(z)\|^2_{G_{\omega}^{\dagger}}
\end{align*}

To finish the proof we need to bound $\vert \max_{z' \in \cC} \epsilon(z')\vert$. Based on the definition of $\epsilon(z')$ we have
\begin{align*}
\vert \epsilon(z') \vert & \leq \vert q_r(z')-\rewardq{\pi}(z') \vert + \vert \rewardq{\pi}(z')- \langle \phi(z'), \theta_r^* \rangle \vert \\ 
& \leq \vert q_r(z)-\rewardq{\pi}(z) \vert + \epsilon_b \\
& \leq \frac{1}{1 - \gamma} \,   \sqrt{\frac{\log(2  /\delta)}{2 m}} + \epsilon_b
\end{align*}
where the second inequality is due to function approximation error (\cref{assum:linear-realizability}) and the last inequality comes from Hoeffding’s inequality. Specifically, since the $m$ trajectories are independent, and the action-value functions lie in the $[0, \nicefrac{1}{1- \gamma}]$ range, we use Hoeffding's inequality to conclude that the sampling error for each $z \in \cC$ can be upper-bounded by $\frac{1}{1 - \gamma} \sqrt{\frac{\log(2/\delta)}{2 m}}$. Since we desire uniform control over all states and actions in $\cC$, by union bound, with probability $1 - \delta$, $|q_r(z) - \rewardq{\pi}(z)| \leq \frac{1}{1 - \gamma} \sqrt{\frac{\log(2 |\cC|/\delta)}{2 m}}$. Putting everything together we get the result. 
\end{proof}

\myquote{\begin{thmbox}
\cblfa*
\end{thmbox} 
}
\begin{proof}
The proof is similar to the proof of~\cref{cor:cb} but with a different $\tvarepsilon$.  
\end{proof}

\clearpage
\section{Additional Implementation Details}
\label{app:exp-details}
In~\cref{app:practical-cbp-algo}, we describe a more practical variant of \Alg/ and in~\cref{app:g-experimental-algo}, we describe the offline G-experimental design procedure required to form the coreset $\cC$ for \Alg/. Details about the synthetic tabular environment are presented in~\cref{app:des-tabular-env} whereas~\cref{app:hyperparameters} details the hyperparameters used across the different experiments. 

\subsection{Practical Coin-Betting Politex algorithm}
\label{app:practical-cbp-algo}
We present the practical version of CBP which uses a parameter $\alpha_\lambda$ in \cref{alg:cbp-practical}.
\begin{algorithm}[!ht]
\LinesNumbered
\caption{Practical Coin-Betting Politex}
\label{alg:cbp-practical}
    \textbf{Input}: $\alpha_\lambda>0$ (parameter), $\pi_0$ (policy initialization), $\lambda_0$ (dual variable initialization), $m$ (Number of trajectories), $T$ (Number of iterations), Feature map $\Phi$. \vspace{1ex} \\
    
    \textbf{Initialize}: $L_0=0$\\
    
    Compute coreset $\cC$ and distribution $\omega$  \vspace{1ex} \\
    
    Solve the unconstrained problem $\max_{\pi} \consthat{\pi}$ to estimate $\zeta$ in~\cref{lemma:sd} and set $U = \frac{2}{\zeta (1 - \gamma)}$.  \vspace{1ex}
    
    \For{$t \leftarrow 0$  \KwTo  $T-1$}{
    For every $(s,a) \in \cC$, use $m$ trajectories starting from $(s,a)$ using policy $\pit$ and estimate the action-value functions $q_r(s,a)$ and $q_c(s,a)$.  \vspace{1ex}
    
    Compute and store $\thetahat_r^{\pit}$ and $\thetahat_c^{\pit}$ using~\cref{eq:lspg}.  \\

    \For{every $s$ encountered in the trajectory generated by $\pit$, and for every $a$}{\vspace{1ex}
    Compute 
    $\rewardqhat{t}(s,a)  = \langle \thetahat^{\pit}_{r}, \phi(s,a) \rangle$; $\constqhat{t}(s,a) = \langle \thetahat^{\pit}_{c}, \phi(s,a) \rangle$ and $\lagqhat{t}(s,a)  = \rewardqhat{t}(s,a) + \lambda_{t} \, \constqhat{t}(s,a)$. \vspace{1ex}
    
    Update policy,
    \begin{align*}
    \lagahat{t}(s,a) &= \frac{1 - \gamma}{1 + U} \, \left[\lagqhat{t}(s,a) - \left \langle \lagqhat{t}(s,\cdot), \pit(\cdot|s) \right \rangle \right]\\
    \lagatilde{t}(s,a) &= \lagahat{t}(s,a) \, \cI\{w_t(s,a) > 0\} + [\lagahat{t}(s,a)]_{+} \, \cI\{w_t(s,a) \leq 0\}\\
    w_{t+1}(s,a) &= \frac{\sum_{i=0}^{t} \lagatilde{i}(s,a)}{(t+1) + T/2} \left(1 + \sum_{i=0}^{t} \lagatilde{i}(s,a) \, w_{i}(s,a) \right)\\
    \pitt(a|s) &= 
    \begin{cases}
        \pi_0(a|s), \quad \text{if } \sum_{a}{\pi_0(a|s) \, [w_{t+1}(s,a)]_{+}} = 0 \\
        \frac{\pi_0(a|s) \, [w_{t+1}(s,a)]_{+}}{\sum_{a}{\pi_0(a|s) \, [w_{t+1}(s,a)]_{+}}}, \quad \text{otherwise}.
    \end{cases}    
    \end{align*}
    }
    Update dual variable,
    \begin{align*}
    \hat V_c^t(\rho)& = \langle \rho(\cdot), \langle \constqhat{t}(s,\cdot), \pit(\cdot|s) \rangle\rangle\\
    g_t &= b - \hat V_c^t(\rho)\\
    L_t &= \max(L_{t-1}, |g_t|)\\
    \lambda_{t+1} &= \lambda_0 + \frac{\sum_{i=0}^{t} g_i}{L_t \max(\sum_{i=0}^t |g_i| + L_t, \alpha_\lambda L_t)}\Big( L_t + \sum_{i=0}^t [(\lambda_i - \lambda_0)g_i]_{+} \Big)
    \end{align*}
    }
\end{algorithm}

\subsection{Offline G-Experimental Design to build coreset $\cC$}
\label{app:g-experimental-algo}
We use offline G-experimental design to form the coreset in Line~$2$ of~\cref{alg:cbp}. In particular, we use the greedy iterative algorithm in~\cref{alg:g-design-coreset-algo} to build $\cC$: in iteration $\tau$, go through all the states and actions adding the $(s,a)$ pair (to $\cC$) with the highest marginal gain computed as $\textsl{g}_{\tau}(s,a) := \indnorm{\phi(s,a)}{G_{\tau}^{\dagger}}$. Here $G_{\tau}$ is the Gram matrix formed by the features of the $(s,a)$ pairs present in $\cC$ at iteration $\tau$. For a specified input $\epsilon' > 0$, the algorithm terminates at iteration $\Tau$ when $\max_{(s,a)} \textsl{g}_{\Tau}(s,a) \leq \epsilon'$. Hence, the algorithm directly controls $\sup_{(s,a)} \indnorm{\phi(s,a)}{G_{\omega}^{\dagger}} \leq \epsilon'$ in~\cref{lemma:lspe-extrapolation}, and hence controls $\epss$ in practice. However, this procedure does not have a guarantee on how large $\vert \cC \vert$ can be. In practice, we set $\epsilon'$ such that $\vert \cC \vert = O(d)$. Although we only consider forming the coreset in an offline manner that involves iterating through all $SA$ state-action pairs, efficient online variants forming the coreset while running the algorithm have been developed recently~\citep{li2021sample}. Such techniques are beyond the scope of this paper and we plan to explore them in future work.

\begin{algorithm}[!h]
\LinesNumbered
\caption{Coreset $\cC$ formation using G-experimental design}
\label{alg:g-design-coreset-algo}
    \textbf{Input}: $\Phi$ (Feature map), $\epsilon'>0$ (tolerance parameter), $\nu = 1$ (default value). \vspace{1ex} \\
    
    \textbf{Initialize}: $G^{\dagger} = \frac{1}{\nu} \cI_{d}$, $\cC = \emptyset$, $g_{max} = \infty$ (maximum marginal gain).\\
    \While{$g_{max} \leq \epsilon'$}{
        $g_{max} = 0$\\
        \For{ $\forall (s,a) \in (\cS \times \cA)$}{
            Compute $g(s,a) = \sqrt{\phi(s,a)^{\transpose} G^{\dagger} \phi(s,a)}$ \quad [marginal gain]\\
            \If{$g_{max} < g(s,a)$}{
            $(s^*, a^*) = (s,a)$\\
            $g_{max} = g(s,a)$
            }
        }
        $\cC = \cC \cup \{(s^*, a^*)\}$ \\
        $G^{\dagger} = G^{\dagger} - \frac{G^{\dagger} \phi(s^*, a^*) \phi(s^*, a^*)^{\transpose} G^{\dagger}}{1 + \phi(s^*, a^*)^{\transpose} G^{\dagger} \phi(s^*, a^*)}$ \quad [Sherman-Morrison to compute $\left(G + \phi(s^*, a^*) \,\phi(s^*, a^*)^{\transpose}\right)^{\dagger}$]
        
    }
\end{algorithm}

\clearpage
\subsection{Synthetic Tabular Environment}
\label{app:des-tabular-env}
\begin{figure}[!ht]
    \centering
    \includegraphics[scale=0.3]{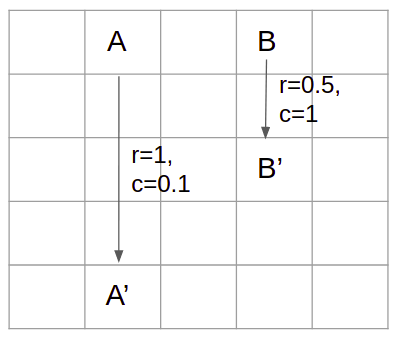}
    \caption{\textbf{Tabular environment:} A $5X5$ gridworld environment where all actions results in reward(r) and constraint reward(c) as $0$, except special states denoted by $A$ and $B$. All four actions in states $(A,B)$ transitions the agent to states $(A', B')$ and results in reward as $(1, 0.5)$ and constraint rewards as $(0.1, 1)$ respectively. The remaining transitions incur zero reward and zero constrain reward.}
    \label{fig:gridworld}
\end{figure}

In \cref{fig:gridworld}, we show the synthetic tabular environment which is modified from Example 3.5 \citep{sutton2018reinforcement} to add the constraint rewards.

\subsection{Hyper-parameters}
\label{app:hyperparameters}
\begin{table}[!h]
\caption[]{\textbf{Hyperparameters for tabular setting:} Shows the hyperparameters for different algorithms CBP, GDA and CRPO for experiments in~\cref{app:expaddtab}.}
\label{tab:hyperparameter-tabular}
\centering
\begin{tabular}{|c|c|c|c|}
\hline
Experiments          & \textbf{CBP}       & \textbf{GDA}                                                                     & \textbf{CRPO}     \\ \hline
\textit{Model-based} & $\alpha_\lambda=8$ & \begin{tabular}[c]{@{}c@{}}$\alpha_\pi=1.0$,\\ $\alpha_\lambda=0.1$\end{tabular} & $\alpha_\pi=0.75,\eta=0.0$ \\ \hline
\textit{Model-free}  & $\alpha_\lambda=8$ & \begin{tabular}[c]{@{}c@{}}$\alpha_\pi=1.0$,\\ $\alpha_\lambda=0.1$\end{tabular} & $\alpha_\pi=0.75,\eta=0.0$ \\ \hline
\end{tabular}
\end{table}
\begin{table}[!h]
\caption[]{\textbf{Hyperparameters for LFA setting with sampling :} Shows the hyperparamters used for different $d$ dimension features for CBP, GDA, CRPO with fixed number of samples for $\hat Q$ approximations (for gridworld experiments in \cref{app:lfa-exp}).}
\centering
\begin{tabular}{|l|l|l|l|}
\hline
\textbf{Algorithms} & $d=40$                  & $d=56$                  & $d=80$                 \\ \hline
\textbf{CBP}        & $\alpha_\lambda=0.25$ & $\alpha_\lambda=0.25$ & $\alpha_\lambda=0.1$ \\ \hline
\textbf{GDA} &
  \begin{tabular}[c]{@{}l@{}}$\alpha_{\pi}=1.0$,\\ $\alpha_{\lambda}=0.1$\end{tabular} &
  \begin{tabular}[c]{@{}l@{}}$\alpha_\pi=1.0$,\\ $\alpha_\lambda=1.0$\end{tabular} &
  \begin{tabular}[c]{@{}l@{}}$\alpha_\pi=1.0$,\\ $\alpha_\lambda=0.1$\end{tabular} \\ \hline
\textbf{CRPO}       & $\alpha_\pi=0.75$     & $\alpha_\pi=0.75$     & $\alpha_\pi=0.75$    \\ \hline
\end{tabular}
\label{tab:hyper_lfa_sampling_tc}
\end{table}

\begin{table}[!h]
\caption[]{\textbf{Hyperparameters for Cartpole environment}: Shows the best hyperparameter for different values of entropy regularization coefficient $\nu$ and different algorithms namely CBP, GDA and CRPO (for experiments in \cref{app:lfa-exp}).}
\centering
\begin{tabular}{|c|c|c|c|c|}
\hline
\textit{Algorithms} & \boldsymbol{$\nu=0$}     & \boldsymbol{$\nu=0.1$}   & \boldsymbol{$\nu=0.01$}  & \boldsymbol{$\nu=0.001$}          \\ \hline
CBP                 & $\alpha_\lambda=0.1$ & $\alpha_\lambda=0.1$ & $\alpha_\lambda=5.0$ & $\alpha_\lambda=0.5$ \\ \hline
GDA &
  \begin{tabular}[c]{@{}c@{}}$\alpha_\pi=0.01$,\\ $\alpha_\lambda=0.001$\end{tabular} &
  \begin{tabular}[c]{@{}c@{}}$\alpha_\pi=0.001$,\\ $\alpha_\lambda=0.001$\end{tabular} &
  \begin{tabular}[c]{@{}c@{}}$\alpha_\pi=0.1$,\\ $\alpha_\lambda=0.1$\end{tabular} &
  \begin{tabular}[c]{@{}c@{}}$\alpha_\pi=0.01$,\\ $\alpha_\lambda=0.0001$\end{tabular} \\ \hline
\textit{CRPO} &
  \begin{tabular}[c]{@{}c@{}}$\alpha_\pi=0.1$,\\ $\eta=10$\end{tabular} &
  \begin{tabular}[c]{@{}c@{}}$\alpha_\pi=0.1$,\\ $\eta=0$\end{tabular} &
  \begin{tabular}[c]{@{}c@{}}$\alpha_\pi=0.5$,\\ $\eta=10$\end{tabular} &
  \begin{tabular}[c]{@{}c@{}}$\alpha_\pi=0.5$,\\ $\eta=0$\end{tabular} \\ \hline
\end{tabular}
\label{tab:hyper_lfa_cartpole}
\end{table}

\clearpage
\clearpage
\section{Additional Experimental Results}
\label{app:experiments}

\subsection{Tabular Setting}
\label{app:expaddtab}
\begin{figure}[!ht]
        \centering
		\begin{subfigure}[b]{0.25\linewidth}
			\centering
			\captionsetup{justification=centering}
			\includegraphics[width=0.9\textwidth]{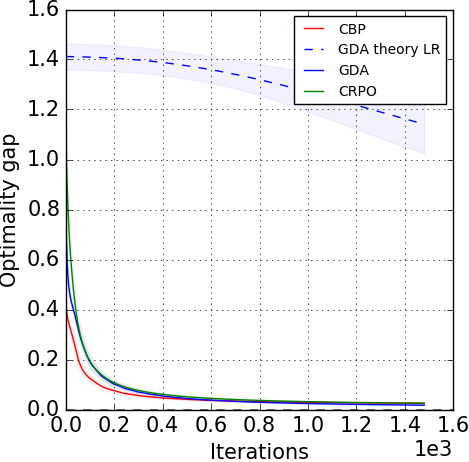}
			\caption[]{\small{OG}}
		\end{subfigure}
		\begin{subfigure}[b]{0.25\linewidth}
			\centering
			\captionsetup{justification=centering}
			\includegraphics[width=0.9\textwidth]{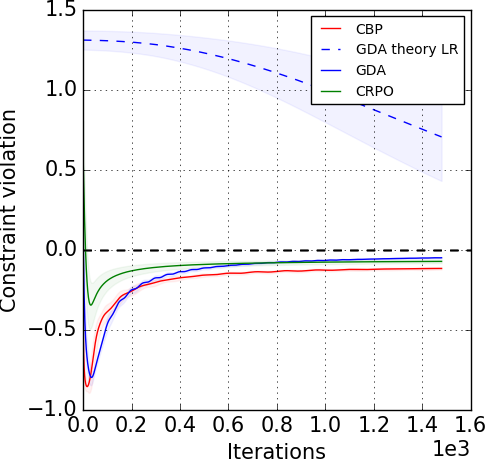}
			\caption[]{\small{CV}}
		\end{subfigure}
		\caption[]
        {\textbf{Model-based in tabular case:} OG and CV for CB, GDA and CRPO with best hyperparameters. The results are averaged over $5$ runs with $95\%$ confidence interval. We assume that we have access to true CMDP. The best hyperparameter has the least OG and satisfies the condition $CV \in[-0.25,0]$. We also show the performance of baseline GDA with theoretical $\alpha_\pi=\sqrt{\frac{2 \log |A|}{T}}\frac{1-\gamma}{1+U}$ and $\alpha_\lambda = \frac{U(1-\gamma)}{\sqrt{T}}$. Here, $U=\frac{2}{\zeta(1-\gamma)}$. This is shown in blue dashed line.}
        \label{fig:CB_GDA_best_param_MB}
\end{figure}

\paragraph{Model-based setting:} In \cref{fig:CB_GDA_best_param_MB}, we demonstrate performance -- optimality gap (OG) and constraint violation (CV) -- with best hyperparameters for three algorithms namely, CBP, GDA and CRPO. In addition, we show the performance of GDA with theoretical learning rates of $\pi$ and $\lambda$ to focus on the importance of tuning GDA's hyperparameter for practical purpose. We observe OG converges to zero quickly for our CBP as compared to GDA and CRPO with constraint satisfaction (when $CV\leq0$). The ideal performance metric is when both OG and CV converges to $0$ value. Refer \cref{tab:hyperparameter-tabular} for best values of hyperparameter. We used the following ranges of hyperparameters. For CBP, $\alpha_\lambda=\{1,2,5,8,15,50,100,300,500\}$. The hyperparameter of GDA varied as $\alpha_\pi = \{0.001, 0.01, 0.1, 1.0\}$ (learning rate policy) and $\alpha_\lambda=\{0.0001, 0.001, 0.01, 0.1, 1.0\}$ (learning rate dual variable). For CRPO, the learning rate of policy varied as $\alpha_\pi=\{0.001, 0.01, 0.05, 0.1, 0.5, 0.75\}$ and tolerance parameter $\eta = \{0, 0.25\}$. 
\label{app:tab-exp}

\begin{figure}[!ht]
\centering
		\begin{subfigure}[b]{0.25\linewidth}
			\centering
			\captionsetup{justification=centering}
			\includegraphics[width=0.9\textwidth]{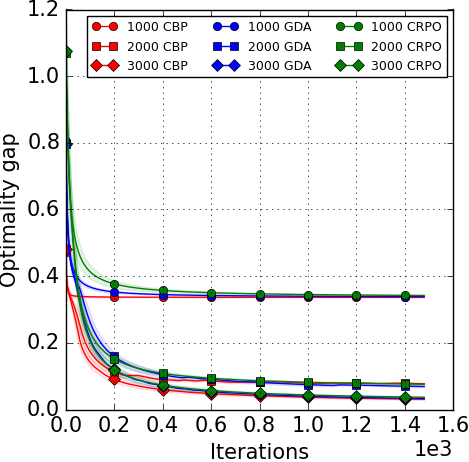}
			\caption[]{\small{OG}}
		\end{subfigure}
		\begin{subfigure}[b]{0.25\linewidth}
			\centering
			\captionsetup{justification=centering}
			\includegraphics[width=0.9\textwidth]{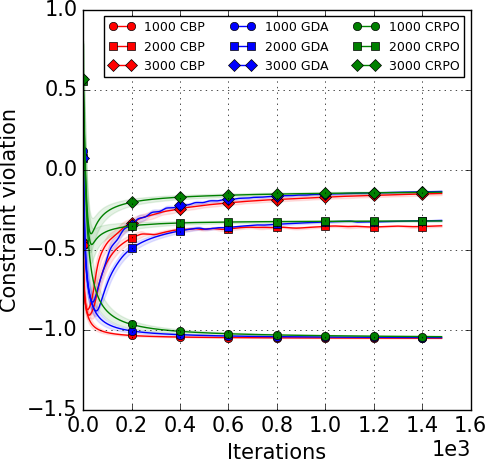}
			\caption[]{\small{CV}}
		\end{subfigure}
		\caption[]
        {\textbf{Model-free tabular case:} We don't have access to true CMDP model here. We vary the $\text{number of samples}=\{1000, 2000, 3000\}$ for $Q$ value estimation to observe the effect on the performance. The results are averaged over $5$ runs. The performance improves with increase in samples. CBP consistently performs better than the baselines GDA and CRPO.}
        \label{fig:CB_GDA_MF_TDSampling}
\end{figure}

\begin{figure}[!ht]
        \centering
		\begin{subfigure}[b]{0.25\linewidth}
			\centering
			\captionsetup{justification=centering}
			\includegraphics[width=0.9\textwidth]{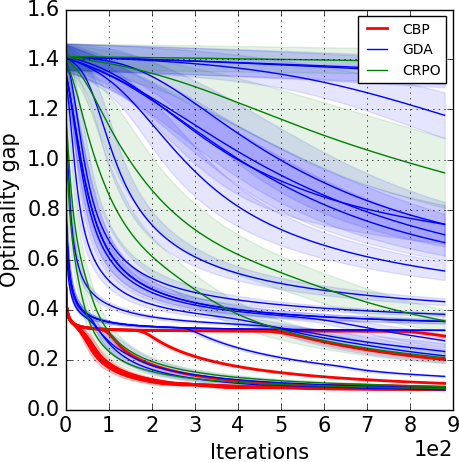}
			\caption[]{\small{OG}}
		\end{subfigure}
		\begin{subfigure}[b]{0.25\linewidth}
			\centering
			\captionsetup{justification=centering}
			\includegraphics[width=0.9\textwidth]{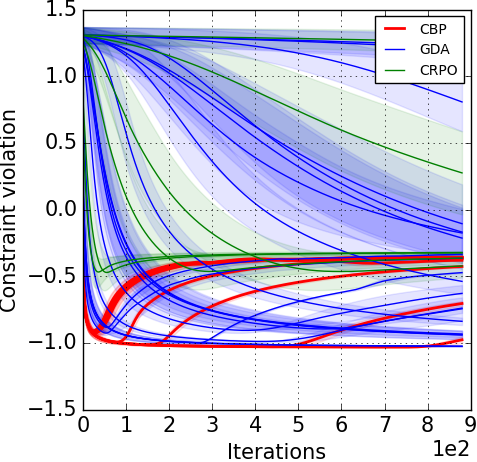}
			\caption[]{\small{CV}}
		\end{subfigure}
		\caption[]
        {\textbf{Sensitivity to hyperparameters in model-free gridworld environment:} Performance with different hyperparameters for CB, GDA and CRPO. The results are averaged over $5$ runs with $95\%$ confidence interval. We used $2000$ samples for $\forall (s,a) \in \cS \times \cA$ to estimate $Q$ values for both reward and cost. The results are demonstrated on one-hot features (with no feature overlap). The hyperparameters for CB are $\alpha_\lambda=\{1, 2, 5, 8, 15, 50, 100, 300, 500\}$. The hyperparameter for GDA are $\alpha_\pi = \{0.001, 0.01, 0.1, 1.0\}$ (learning rates for policy) and $\alpha_\lambda=\{0.0001, 0.001, 0.01, 0.1, 1.0\}$ (learning rate for dual variable). The hyperparameter for CRPO are $\alpha_\pi= \{0.001, 0.01, 0.05, 0.1, 0.5, 0.75\}$. We use $\eta=0.0$ for CRPO. The key observation is that CBP is robust against the variations in hyperparameters with a smaller variance in performance against multiple runs.}
        \label{fig:MF_sampling_1_hot_hyperparam_sensitivity}
\end{figure}

\begin{figure}[!ht]
        \centering
		\begin{subfigure}[b]{0.25\linewidth}
			\centering
			\captionsetup{justification=centering}
			\includegraphics[width=0.9\textwidth]{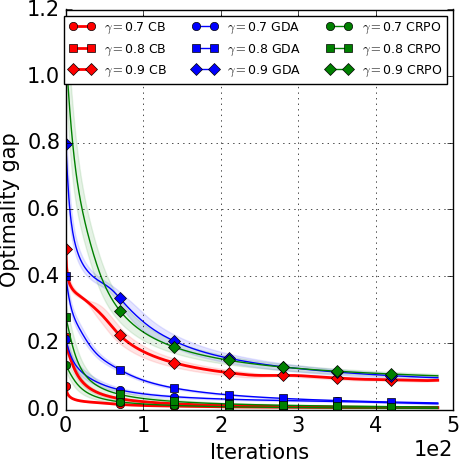}
			\caption[]{\small{OG}}
		\end{subfigure}
		\begin{subfigure}[b]{0.25\linewidth}
			\centering
			\captionsetup{justification=centering}
			\includegraphics[width=0.9\textwidth]{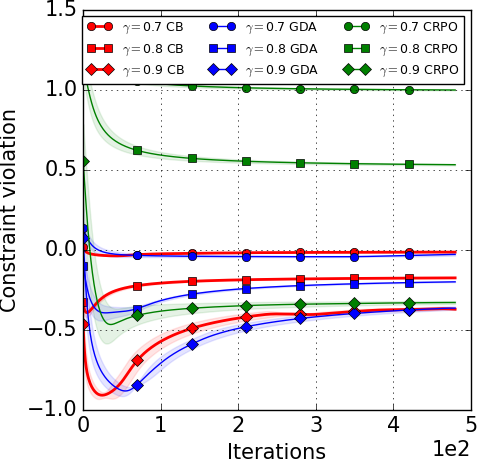}
			\caption[]{\small{CV}}
		\end{subfigure}
		\caption[]
        {\textbf{Environment misspecification in model-free gridworld by varying discount factor $\gamma$:}  We only introduce sampling error by estimating $Q$ function with $2000$ samples for all $(s,a)$ pair for all three algorithms (no feature overlap). We vary the discount factor $\gamma=\{0.7, 0.8\}$ to observe the effects of environment misspecification on CBP, GDA and CRPO. We keep the hyperparameters fixed for all the algorithms, similar to the one for original CMDP with $\gamma=0.9$. The results are averaged over $5$ runs with $95\%$ confidence interval.The hyperparameters used are reported in \cref{tab:hyperparameter-tabular}. We observe that CRPO does not even satisfy constraint for case when $\gamma=0.8$ ($CV>0$). Further, our CBP converges consistently faster than the baselines.}
        \label{fig:MF_sampling_1_hot_gamma}
\end{figure}

\paragraph{Model-free setting:} Here, we test the performance of algorithms in the model-free setting (don't have access to true CMDP model). We use TD(0) based sampling approach \citep{sutton1988learning} to estimate the $Q$ action-value function. We sample data for all $(s,a) \in \cS \times \cA$. In \cref{fig:CB_GDA_MF_TDSampling}, we observe the effect on performance by varying the number of samples for $Q$ action-value estimation. Here, we consider one-hot encoded features (no overlapping of features). We observe that CBP consistently converges faster than its counterpart in the sampling-based approach. Further, it also matches the expectation that the performance improves with the increase in number of samples. In~\cref{fig:MF_sampling_1_hot_hyperparam_sensitivity,fig:MF_sampling_1_hot_gamma}, we show the robustness of CBP with hyperparameter sensitivity and environment misspecification respectively. The hyperparameters used for these experiments are presented in \cref{tab:hyperparameter-tabular} in \cref{app:hyperparameters}.

\subsection{Linear setting}
\label{app:lfa-exp}
\begin{figure}[!ht]
        \centering
		\begin{subfigure}[b]{0.25\linewidth}
			\centering
			\captionsetup{justification=centering}
			\includegraphics[width=0.9\textwidth]{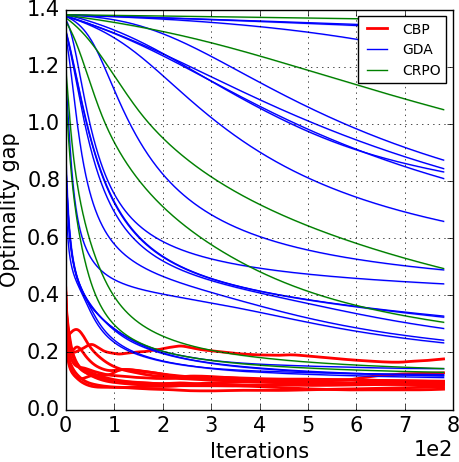}
			\caption[]{\small{OG ($d=40$)}}
		\end{subfigure}
		\begin{subfigure}[b]{0.25\linewidth}
			\centering
			\captionsetup{justification=centering}
			\includegraphics[width=0.9\textwidth]{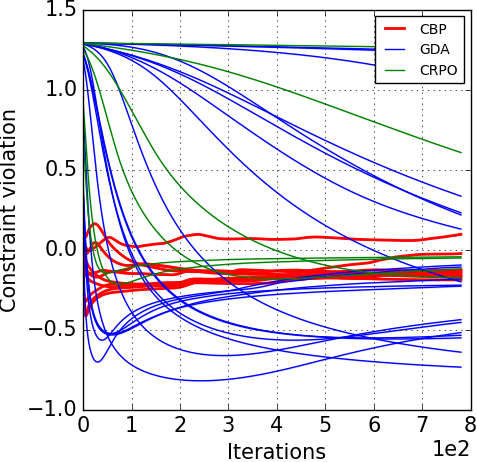}
			\caption[]{\small{CV ($d=40$)}}
		\end{subfigure}
		
		\begin{subfigure}[b]{0.25\linewidth}
			\centering
			\captionsetup{justification=centering}
			\includegraphics[width=0.9\textwidth]{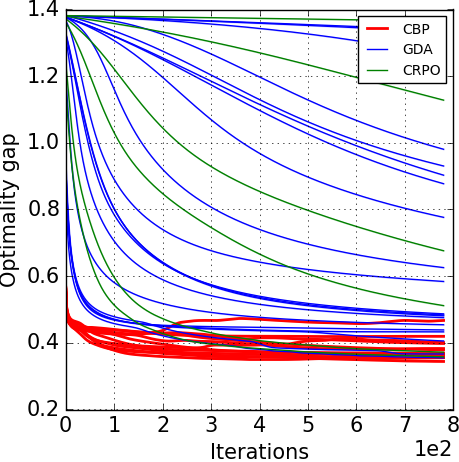}
			\caption[]{\small{OG ($d=56$)}}
		\end{subfigure}
		\begin{subfigure}[b]{0.25\linewidth}
			\centering
			\captionsetup{justification=centering}
			\includegraphics[width=0.9\textwidth]{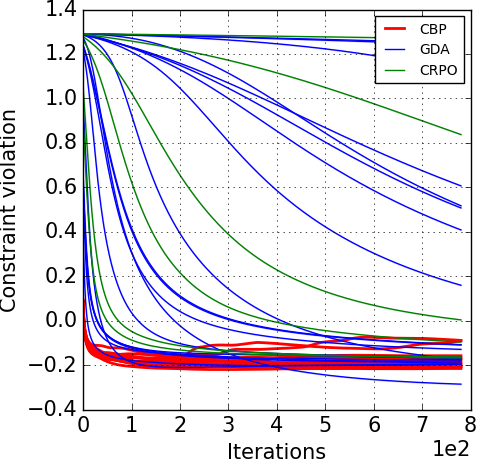}
			\caption[]{\small{CV ($d=56$)}}
		\end{subfigure}
		
		\begin{subfigure}[b]{0.25\linewidth}
			\centering
			\captionsetup{justification=centering}
			\includegraphics[width=0.9\textwidth]{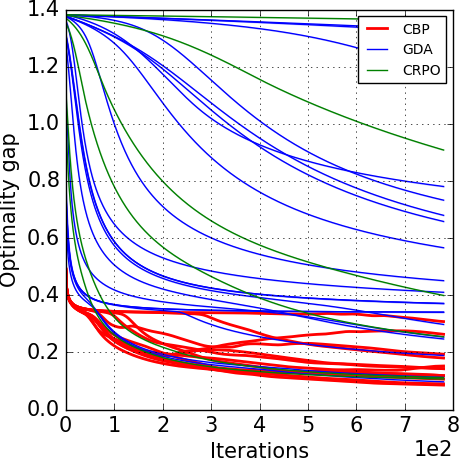}
			\caption[]{\small{OG ($d=80$) }}
		\end{subfigure}
		\begin{subfigure}[b]{0.25\linewidth}
			\centering
			\captionsetup{justification=centering}
			\includegraphics[width=0.9\textwidth]{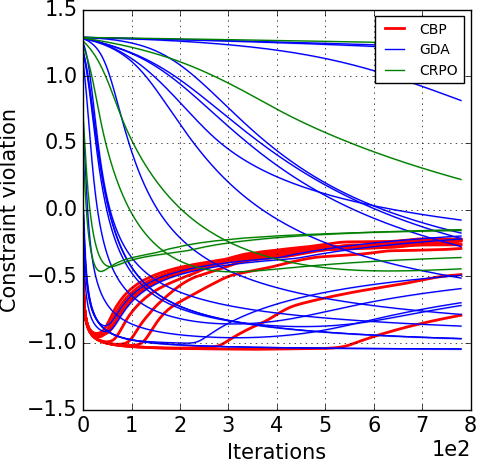}
			\caption[]{\small{CV ($d=80$)}}
		\end{subfigure}
		\caption[]
        {\textbf{Linear function approximation in gridworld environment:} We approximate the $Q$ value using LSTDQ with $300$ samples for each $(s,a)$ pair. The dimension of the features (denoted by $d$) are varied to observe the sensitivity to a range of hyperparameter values. We kept all the other parameters fixed. We use (a,b) $d=40$, (b,c) $d=56$, (e,f) $d=80$ dimension features respectively. \cAlg/ is consistently robust to variation in the hyperparameters for different dimensions as compared to baselines \cGDA/ and \cCRPO/.}
        \label{fig:lfa-gridworld-hyperparmeter-sensitivity}
\end{figure}

\paragraph{Gridworld environment:} We use $5\times5$ gridworld environment as show in \cref{fig:gridworld}. Tile coding is used to learn the feature representation for every $(s,a)$ pair in the environment. Number of tilings used are $1$ and we vary the tiling size to change the dimension of the features (feature overlap for multiple $(s,a)$ pairs). In \cref{fig:lfa-gridworld-hyperparmeter-sensitivity} we show hyperparameter sensitivity on performance for all the three algorithms with different $d$ dimension of features. The values of all other parameters were kept fixed. Similar observation holds here, CBP is robust to varying values of hyperparameters. The range of hyperparameter is similar to one in \cref{fig:MF_sampling_1_hot_hyperparam_sensitivity}.

\begin{figure}[!ht]
\centering
		\begin{subfigure}[b]{0.25\linewidth}
			\centering
			\captionsetup{justification=centering}
			\includegraphics[width=0.9\textwidth]{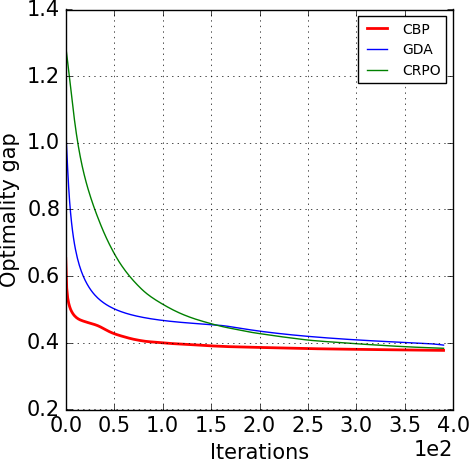}
			\caption[]{\small{OG}}
		\end{subfigure}
		\begin{subfigure}[b]{0.25\linewidth}
			\centering
			\captionsetup{justification=centering}
			\includegraphics[width=0.9\textwidth]{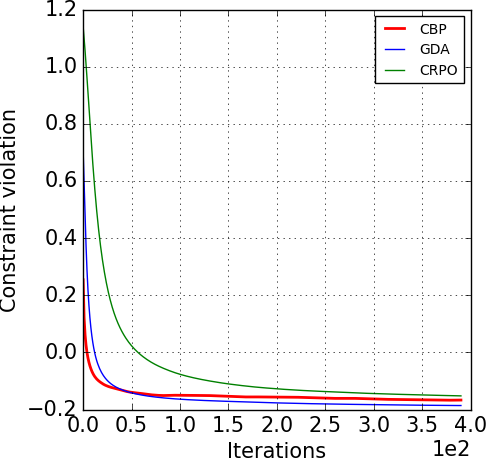}
			\caption[]{\small{CV}}
		\end{subfigure}
		\caption[]
        {\textbf{G-experimental design:} We show the performance with G-experimental design, where subset of $(s,a)\in \cC$ are chosen to learn the  weight vectors of $Q$ function. Here, we are in model-free setting with $d=56$ dimensional features in LFA. We used $300$ samples for $c \in \cC$ to learn the estimate of $Q$ for all three algorithms.}
        \label{fig:g-experimental}
\end{figure}

\paragraph{G-experimental design for gridworld environment:} In~\cref{fig:g-experimental} we show the performance with G-experimental design (\cref{app:g-experimental-algo}). Here subset of $(s,a)\in \cC$ pairs are chosen from a \textit{coreset}. 

 \paragraph{Exploration in continuous state-spaces:} We used G-experimental design for the discrete state-action environment in the previous section. However, such a procedure is difficult to implement for the continuous state-action spaces we consider in this section. In order to achieve enough exploration in practice, similar to~\citet{xu2021crpo}, we use entropy regularization~\citep{geist2019theory,cen2021fast} for the policy updates. Specifically, for a specified regularization parameter $\nu$, our task is to find a sequence of policies $\{\pi_0, \pi_1, \ldots, \pi_{T-1}\}$ that minimize the regularized primal regret,
\begin{align*}
\cR^{p}_{\nu}(\pi^*, T) & := \sum_{t = 0}^{T-1} \sum_{s = 0}^{\cS-1} \big[\langle \piopt(\cdot|s) - \pit(\cdot|s), \rewardqhat{t} + \lambdat \constqhat{t} \rangle + \nu \, d^{\pit}(s) \sum_{a \in \cA} \pi(a|s) \log(\pi(a|s)) \big]. 
\end{align*}
It can be easily seen~\citep{geist2019theory} that the form of the algorithm updates remain the same, but the action-value functions for policy $\pi$ need to be redefined to depend on the ``effective reward'' equal to $r(s,a) - \nu \log \pi(a|s)$. Therefore, the new $\hat Q_l^t$ with exploration is equal to $\hat Q_l^t(s,a) = \hat Q_r^t(s,a) +\lambdat \hat Q_c^t(s,a) - \nu \log \pi_t(a|s)$. 

\paragraph{Cartpole environment}: We added two constraint rewards ($c_1, c_2$) to the classic OpenAI gym Cartpole environment. (1) Cart receives a $c_1=0$ constraint reward value when enters the area $[-2.4, -2.2], [-1.3, -1.1], [1.1, 1.3], [2.2, 2.4]$, else receive $c_1=+1$. (2) When the angle of the cart is less than $4$ degrees receive $c_2=+1$, else everywhere $c_2=0$. Each episode length is no longer than 200.

\begin{figure}[!ht]
        \centering
		\begin{subfigure}[b]{0.25\linewidth}
			\centering
			\captionsetup{justification=centering}
			\includegraphics[width=0.9\textwidth]{Cartpole/5Runs/BReturn_eps_6_e0.0.png}
			\caption[]{\small{Return ($\nu=0$)}}
		\end{subfigure}
		\begin{subfigure}[b]{0.25\linewidth}
			\centering
			\captionsetup{justification=centering}
			\includegraphics[width=0.9\textwidth]{Cartpole/5Runs/BCV1_eps_6_e0.0.png}
			\caption[]{\small{CV 1 ($\nu=0$)}}
		\end{subfigure}
		\begin{subfigure}[b]{0.25\linewidth}
			\centering
			\captionsetup{justification=centering}
			\includegraphics[width=0.9\textwidth]{Cartpole/5Runs/BCV2_eps_6_e0.0.png}
			\caption[]{\small{CV 2 ($\nu=0$)}}
		\end{subfigure}
		
		\begin{subfigure}[b]{0.25\linewidth}
			\centering
			\captionsetup{justification=centering}
			\includegraphics[width=0.9\textwidth]{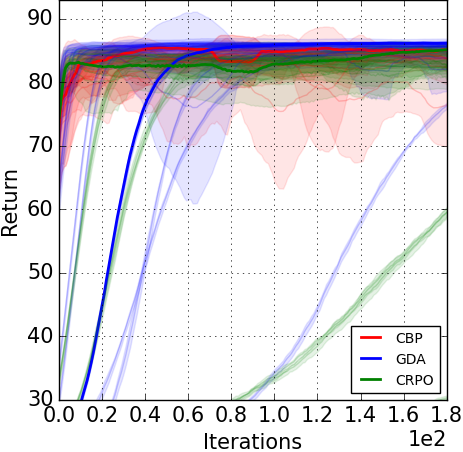}
			\caption[]{\small{Return ($\nu=0.1$)}}
		\end{subfigure}
		\begin{subfigure}[b]{0.25\linewidth}
			\centering
			\captionsetup{justification=centering}
			\includegraphics[width=0.9\textwidth]{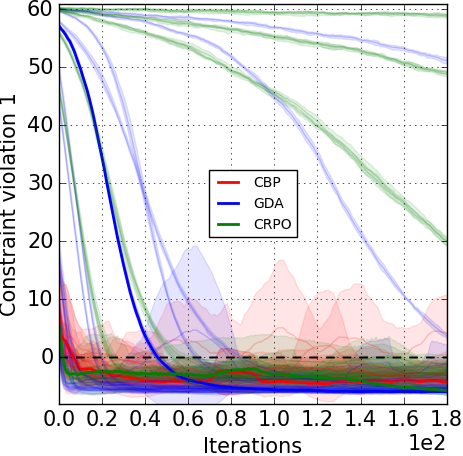}
			\caption[]{\small{CV 1 ($\nu=0.1$)}}
		\end{subfigure}
		\begin{subfigure}[b]{0.25\linewidth}
			\centering
			\captionsetup{justification=centering}
			\includegraphics[width=0.9\textwidth]{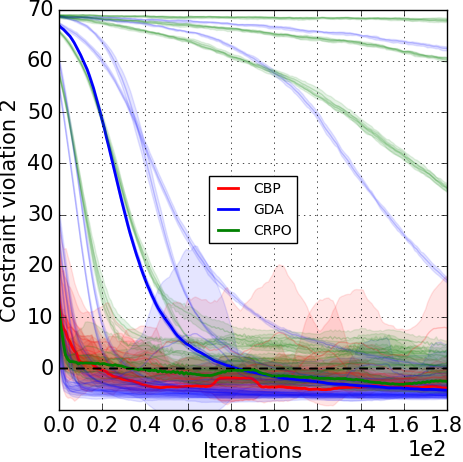}
			\caption[]{\small{CV 2 ($\nu=0.1$)}}
		\end{subfigure}
		
		\begin{subfigure}[b]{0.25\linewidth}
			\centering
			\captionsetup{justification=centering}
			\includegraphics[width=0.9\textwidth]{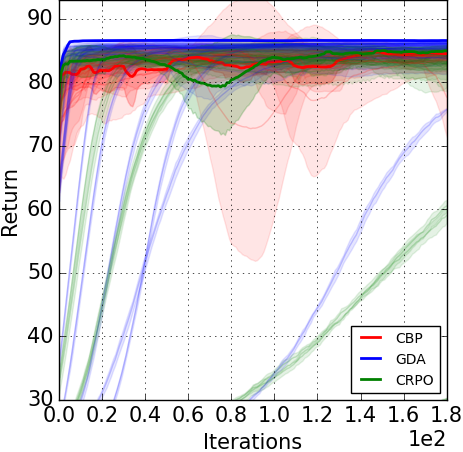}
			\caption[]{\small{Return ($\nu=0.01$)}}
		\end{subfigure}
		\begin{subfigure}[b]{0.25\linewidth}
			\centering
			\captionsetup{justification=centering}
			\includegraphics[width=0.9\textwidth]{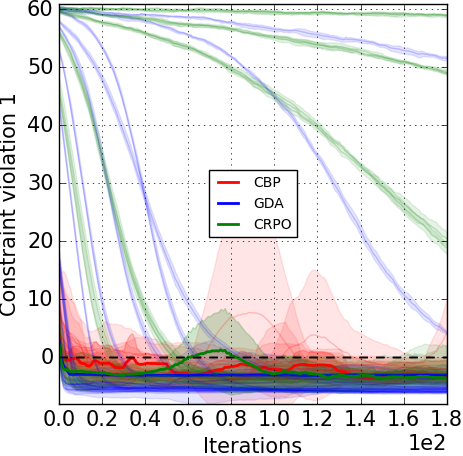}
			\caption[]{\small{CV 1 ($\nu=0.01$)}}
		\end{subfigure}
		\begin{subfigure}[b]{0.25\linewidth}
			\centering
			\captionsetup{justification=centering}
			\includegraphics[width=0.9\textwidth]{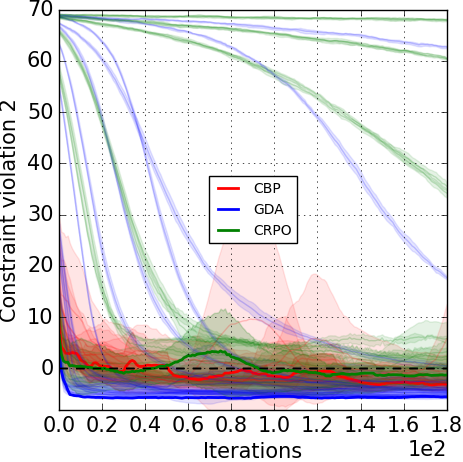}
			\caption[]{\small{CV 2 ($\nu=0.01$)}}
		\end{subfigure}
		
		\begin{subfigure}[b]{0.25\linewidth}
			\centering
			\captionsetup{justification=centering}
			\includegraphics[width=0.9\textwidth]{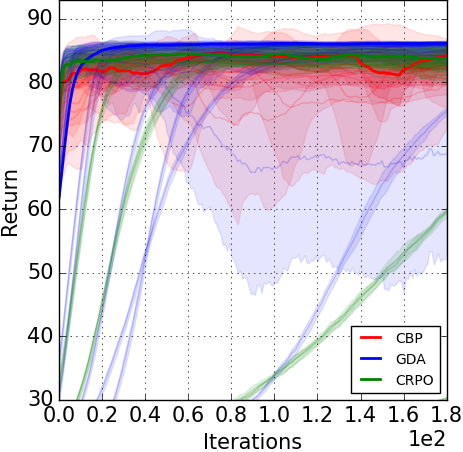}
			\caption[]{\small{Return ($\nu=0.001$)}}
		\end{subfigure}
		\begin{subfigure}[b]{0.25\linewidth}
			\centering
			\captionsetup{justification=centering}
			\includegraphics[width=0.9\textwidth]{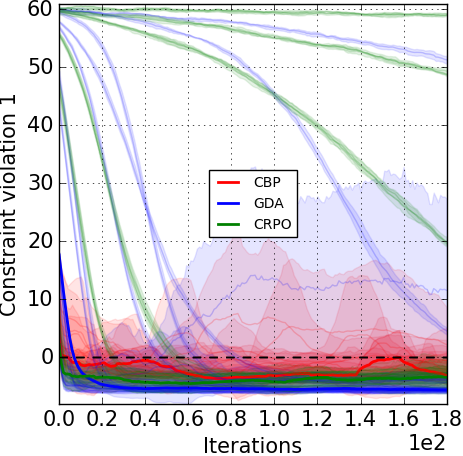}
			\caption[]{\small{CV 1 ($\nu=0.001$)}}
		\end{subfigure}
		\begin{subfigure}[b]{0.25\linewidth}
			\centering
			\captionsetup{justification=centering}
			\includegraphics[width=0.9\textwidth]{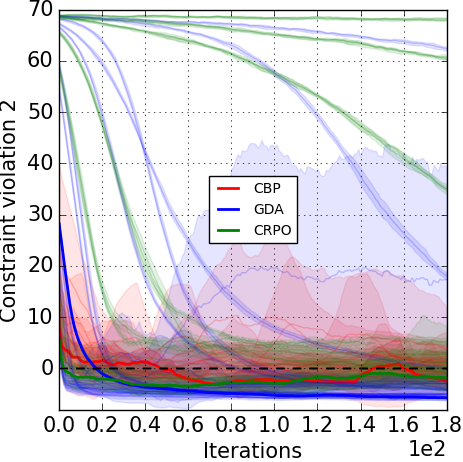}
			\caption[]{\small{CV 2 ($\nu=0.001$)}}
		\end{subfigure}
		\caption[]
        {\textbf{Cartpole environment:} We show the sensitivity to entropy regularization $\nu$ for all three algorithms CBP, GDA and CRPO. The performance is averaged over $5$ runs with $95\%$ confidence interval. Different rows corresponds to different value of $\nu=\{0, 0.1, 0.01, 0.001\}$. Darker lines show the performance with the best hyperparameters. Lighter shade lines show performance with other values of hyperparameters. The range of hyperparameter for \Alg/ is $\alpha_\lambda=\{0.1, 0.5, 5, 50, 250, 500, 750, 1000\}$. For GDA, we vary learning rates of both policy and dual variable as $\{0.1, 0.01, 0.001, 0.0001\}$. For CRPO, we vary $\alpha_\pi=\{0.1, 0.5, 0.01, 0.001, 0.005\}$ and tolerance hyperparameter as $\eta=\{0, 10\}$.}
        \label{fig:cartpole_entropy_sensitivity}
\end{figure}

We used tile coding \citep{sutton2018reinforcement} to discretize the continuous state space of the environment. The dimension of the features is $2^{12}$. We used $8$ number of tilings with each grid size $4\times 4$. For experimenting the effect of adding exploration on the performance, we incorporated the entropy coefficient \citep{haarnoja2018soft,geist2019theory}. We varied the entropy regularizer $\nu = \{0, 0.1, 0.01, 0.001\}$. Refer to \cref{fig:cartpole_entropy_sensitivity} for the experiment with $\nu$ coefficient.

We conducted the experiments with following $\alpha_\lambda$ parameter value of CBP $\{0.1, 0.5, 5, 50, 250, 500, 750, 1000\}$. For GDA, we varied the learning rate of policy $\alpha_\pi = \{0.1, 0.01, 0.001, 0.0001\}$ and learning rate of dual variable $\alpha_\lambda= \{0.1, 0.01, 0.001, 0.0001\}$. For CRPO baseline, the following values of learning rate of policy $\alpha_\pi=\{0.001, 0.005, 0.01, 0.05, 0.1, 0.5\}$ are experimented with. We kept the tolerance parameter $\eta$ of CRPO as $\{0, 10\}$. The best hyperparameters are summarized in \cref{tab:hyper_lfa_cartpole} for the different values of entropy regularizer $\nu$.
\end{document}